%% file: main.tex
\documentclass[10pt,twocolumn,letterpaper]{article}

\usepackage{cvpr}              


\definecolor{cvprblue}{rgb}{0.21,0.49,0.74}
\usepackage[pagebackref,breaklinks,colorlinks,allcolors=cvprblue]{hyperref}
\usepackage{graphicx}
\usepackage{caption}
\usepackage{comment}

\def\paperID{4295}
\def\confName{CVPR}
\def\confYear{2025}

\usepackage[accsupp]{axessibility}

\title{Conditional Balance:\\ Improving Multi-Conditioning Trade-Offs in Image Generation}

\author{Nadav Z. Cohen$^1$, Oron Nir$^{1,2}$, Ariel Shamir$^1$ \\
\small $^1$Reichman University, $^2$Microsoft Corporation \\
\small \href{https://nadavc220.github.io/conditional-balance.github.io/}{\textcolor{teal}{\texttt{\textit{https://nadavc220.github.io/conditional-balance.github.io/}}}}
}

\begin{document}

\twocolumn[{
    \maketitle
    \begin{center}
        \vspace{-1.5em}
        \includegraphics[width=\textwidth]{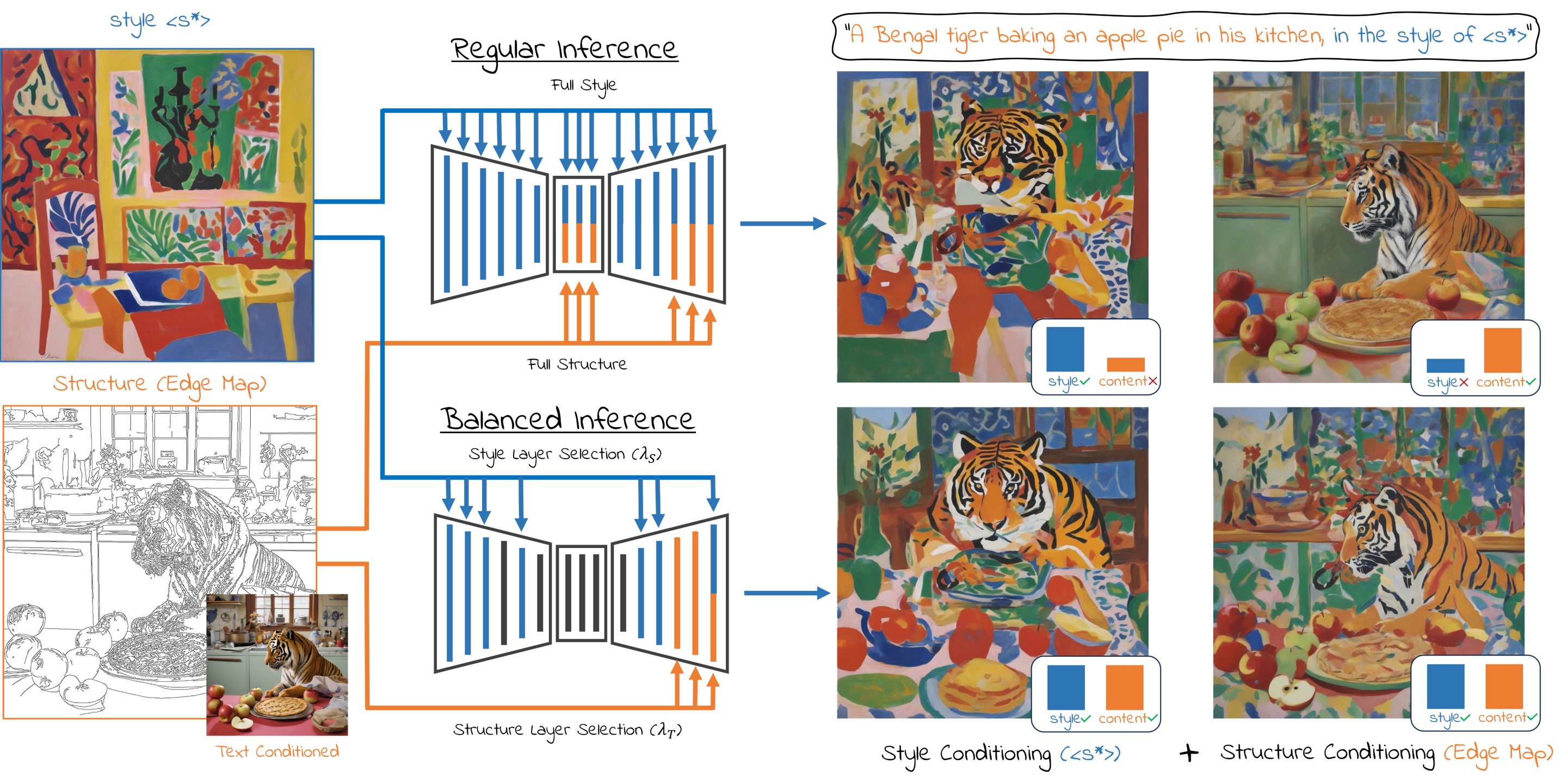}
        \captionof{figure}{\textbf{Balanced Conditioning Image Generation.} \textit{We analyze the sensitivity of model layers to various aspects in conditional inputs. This allows to limit the inputs only to specific layers during inference, thereby balancing the different conditions and preventing content and style from overshadowing each other. As a result, the generative model reduces artifacts and gains artistic freedom when combining complex conditional inputs. As can be seen, by selecting only highly sensitive layers of style $(\lambda_S)$ and structure $(\lambda_T)$ we get better color and texture control and better geometric style control. }} 
        \label{fig:teaser}
        \vspace{0.1em}  
    \end{center}
}]


\input{sec/0_abstract_v1}
\input{sec/1_intro_v2}

\input{sec/2_related_work_v2}

\input{sec/3_tradeoff}

\input{sec/4_method}

\input{sec/5_Analysis}

\input{sec/6_Results}

\input{sec/7_Discussion}


{
    \small
    \bibliographystyle{plain}
    \bibliography{main}
}

\input{sec/X_suppl}

\end{document}

%% file: sec/0_abstract_v1.tex
\begin{abstract}

Balancing content fidelity and artistic style is a pivotal challenge in image generation. While traditional style transfer methods and modern Denoising Diffusion Probabilistic Models (DDPMs) strive to achieve this balance, they often struggle to do so without sacrificing either style, content, or sometimes both. This work addresses this challenge by analyzing the ability of DDPMs to maintain content and style equilibrium. We introduce a novel method to identify sensitivities within the DDPM attention layers, identifying specific layers that correspond to different stylistic aspects. By directing conditional inputs only to these sensitive layers, our approach enables fine-grained control over style and content, significantly reducing issues arising from over-constrained inputs. Our findings demonstrate that this method enhances recent stylization techniques by better aligning style and content, ultimately improving the quality of generated visual content.

\end{abstract}

%% file: sec/1_intro_v2.tex
\section{Introduction}
\label{sec:intro}

To master different aspects of paintings such as color and light, a notable approach practiced by artists involves creating a collection of paintings of the same subject under varying conditions. A well known example of this practice is Claude Monet's series of paintings of Rouen Cathedral~\cite{rouen}. 

Today, modern Denoising Diffusion Probabilistic Models (DDPMs) allow  creating high-quality images of any subject in various styles by iteratively refining random samples. 
To direct these models to output an image with a desirable content at inference, conditional inputs were developed, starting from descriptive text prompts and continuing with images that condition the output to align with content and style information. 
This process, while powerful, tends to lose its conditional constraining ability as the number or complexity of constraints increase. 
As a result, in addition to losing the ability to control the generated content, the model's attempt to satisfy all conditionals leads to more undesirable issues such as image artifacts.

In this work, we investigate the style-content trade-offs of different conditional inputs separately and in combination in generative models.
Our experiments reveal that many issues in conditional generation arise from over-conditioning and the combination of conditionals that were underrepresented during model training.
We analyze a diffusion generation process (using SDXL's architecture~\cite{podell2023sdxlimprovinglatentdiffusion}) and isolate different aspects of its generative capabilities.

Drawing inspiration from Monet's many series of paintings, our analysis uses a collections of images where the content subject is fixed, and a specific stylistic aspect is varied. Using such collections we examine and rank the sensitivity of each layer at each timestep. Later, we direct different conditionals only to specific, sensitive layers, allowing to better balance different conditions (see \cref{fig:teaser}). 
We show that even without pure disentanglement this approach reduces artifacts caused by over-conditioning and enhances the model's overall output quality, stylistic freedom, and consistency without the need of additional training.

We summarize our contributions as follows:
\begin{enumerate}
  \item We develop a novel analysis method inspired by classic art and simple modern statistics which reveals sensitivities in the generation process of diffusion models. 
  \item We analyze SDXL and use our findings to reduce conditional inputs in the generation process. This leads to balance, artifact reduction, and general image quality improvement for text, image, and style conditioning inputs.
  \item Using our findings also allows a flexible way to control content and style aspects of the generated image, leading to a more stable and creative method to generate content.
\end{enumerate}
Our complex-conditioning evaluation set and code are available through our project page.

%% file: sec/2_related_work_v2.tex
\section{Related Work}

\paragraph{Content-Style Applications.}
Seminal style transfer methods~\cite{efros_style, jing2018neuralstyletransferreview, Gatys_2016_CVPR, huang2017adain, johnson2016perceptuallossesrealtimestyle}, explore different ways to transfer a style of a stylistic image to a given ``content image.’' These works demonstrate the natural trade-off between artistic style and content preservation, and explore ways to combine them in a visually pleasing way. Later, inspired by the success of Generative Adversarial Networks~\cite{goodfellow2014generativeadversarialnetworks}, more methods were developed for styled generation~\cite{karras2019stylebasedgeneratorarchitecturegenerative, CycleGAN2017} enabling generating images from noise samples. Although these methods show improvement in combining content and style, they lack flexibility as they can not generate combinations outside of their optimized domain.

\begin{figure*}[t]  
    \centering
    \includegraphics[width=\textwidth]{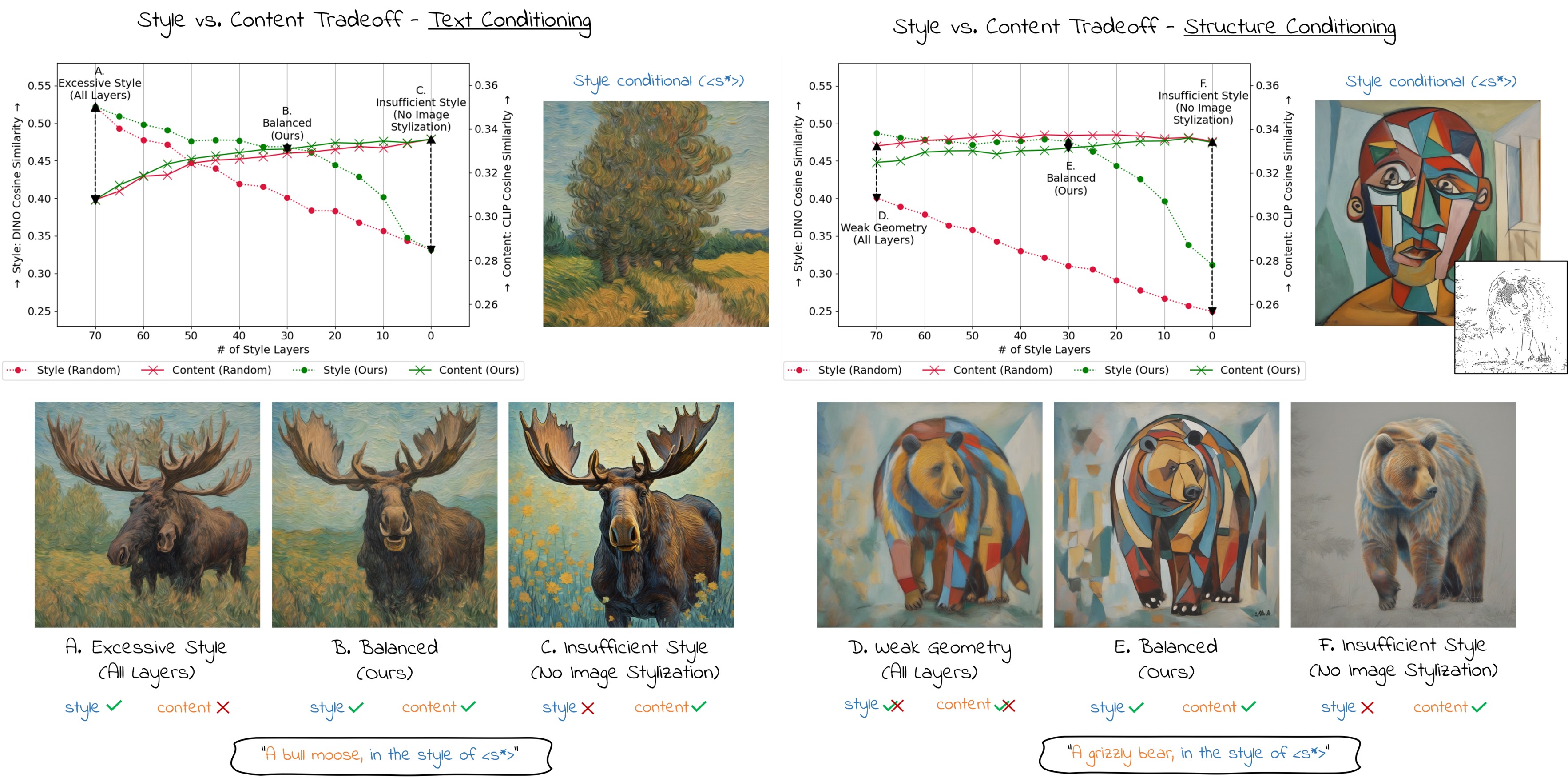}
    \caption{\textbf{Preliminary Experiment Results.} \textit{We present outcomes of selecting sub-set of layers to apply conditioning. We show the content/style tradeoff both with text conditioning (left) and with edge-map conditioning (right). As can be observed in both cases using too much or not enough style conditioning generates an imbalanced output. Selecting layers randomly does not assist much while using our layer reduction strategy enables achieving good quality conditional combinations with less than 50\% of the layers. Note that the content and style graphs are presented on different scales as Clip and Dino similarity values are not aligned.}}
    \label{fig:perlimanery_fig}
\end{figure*}

\paragraph{Diffusion Models and Conditioning.}
DDPMs~\cite{dhariwal2021diffusionmodelsbeatgans, rombach2021highresolution, ramesh2021zeroshottexttoimagegeneration, saharia2022photorealistictexttoimagediffusionmodels, podell2023sdxlimprovinglatentdiffusion} achieved a significant leap in generating novel images conditioned on text prompts by refining over text-image paired datasets~\cite{schuhmann2022laion5bopenlargescaledataset}. While these models combine content and style effectively, their sole reliance on text conditioning comes with shortcomings, such as failure to align to complex text conditions~\cite{Liu2024ImprovingLA, wang2024textanchoredscorecompositiontackling, zhang2024longclip} and an inferior ability to generate content and style combinations which were underrepresented during training.

To overcome these limitations, various conditioning methods were developed, influencing the generated image to resemble information from another image. ControlNet~\cite{zhang2023adding} conditions the output with structure information using image maps such as Canny, depth, and pose. Gal et al.~\cite{gal2022textual}, and others~\cite{gal2023encoderbaseddomaintuningfast, Ruiz_2023_CVPR, arar2024palp}, condition the generated image to preserve unique (personalized) properties of an object.

Image-based style conditioning is a thoroughly researched area~\cite{ye2023ip-adapter, frenkel2024implicit, wang2024instantstyle, jeong2024visual, hertz2023StyleAligned, Chen_2024_CVPR, sohn2023styledroptexttoimagegenerationstyle, cui2024instastyle, rout2024rbmodulation}, usually demonstrated on SDXL~\cite{podell2023sdxlimprovinglatentdiffusion} for its artistic superiority. 
IP-Adapter~\cite{ye2023ip-adapter} and InstantStyle~\cite{wang2024instantstyle} inject the style using a pretrained dedicated Cross-Attention layer~\cite{NIPS2017_3f5ee243} for the conditioning image. B-LoRA~\cite{frenkel2024implicit} and ZipLoRA~\cite{shah2023ZipLoRA} uses LoRA~\cite{hu2022lora} to fine-tune residual weights to match content and style of conditioning images. StyleAligned method~\cite{hertz2023StyleAligned} generates a style image in parallel to generating the output image and injects the style information using AdaIN~\cite{huang2017adain} between Self-Attention layers~\cite{NIPS2017_3f5ee243} while Jeong et al.~\cite{jeong2024visual} extend this idea by replacing the attention feature directly.

Even though recent methods address content and style combinations, they still remain a challenge as methods which excel in style, often sacrificing content fidelity~\cite{hertz2023StyleAligned, jeong2024visual}, while others which preserve content fidelity may often show inaccurate style~\cite{wang2024instantstyle, frenkel2024implicit}.

\paragraph{DDPM Model Analysis.}
Recent works~\cite{zhang2023prospect, frenkel2024implicit, jeong2024visual, wang2024instantstyle} analyze the diffusion process to pinpoint parts responsible for generating various visual aspects. ProSpect~\cite{zhang2023prospect} examine how varying text conditions at different timesteps impact aspects like material, artistic style, and content alignment. Our findings align with theirs, revealing that each timestep serves a unique conditioning function. However, we extend this analysis by investigating the model's internal layers, allowing for the use of multiple conditional inputs without interference. While ProSpect focuses exclusively on text conditioning, our study incorporates both text and image conditioning.
Style conditioning works~\cite{frenkel2024implicit, wang2024instantstyle}, examine internal layers to pinpoint those sensitive to content and style. B-LoRA tests prompt stylization effects in different attention blocks and employs LoRA~\cite{hu2022lora} to fine-tune style sensitive weights for specific images. InstantStyle analyzes SDXL attention layers for content alignment and style sensitivity and concludes the same style-sensitive blocks. 
In contrast, our approach thoroughly evaluates each self-attention layer independently, ranking its sensitivity for each timestep.
Additionally, our method can analyze other attention architectures, like the Joint-Attention layer used by the more recent SD3 model family~\cite{esser2024scalingrectifiedflowtransformers}.
This methodology maximizes style and content conditioning while minimizing interference between layers, thus enhancing stylistic freedom and reducing content artifacts.

\label{sec:related_work}

%% file: sec/3_tradeoff.tex
\section{Content-Style Tradeoff}
\label{sec:tradeoff}


In this section, we explore the issue of over-constrained conditional image generation.
To illustrate the problem, we examine two recent style conditioning methods: StyleAligned~\cite{hertz2023StyleAligned} and B-LoRA~\cite{frenkel2024implicit}. 

\begin{figure*}[t]  
    \centering
    \includegraphics[width=\textwidth]{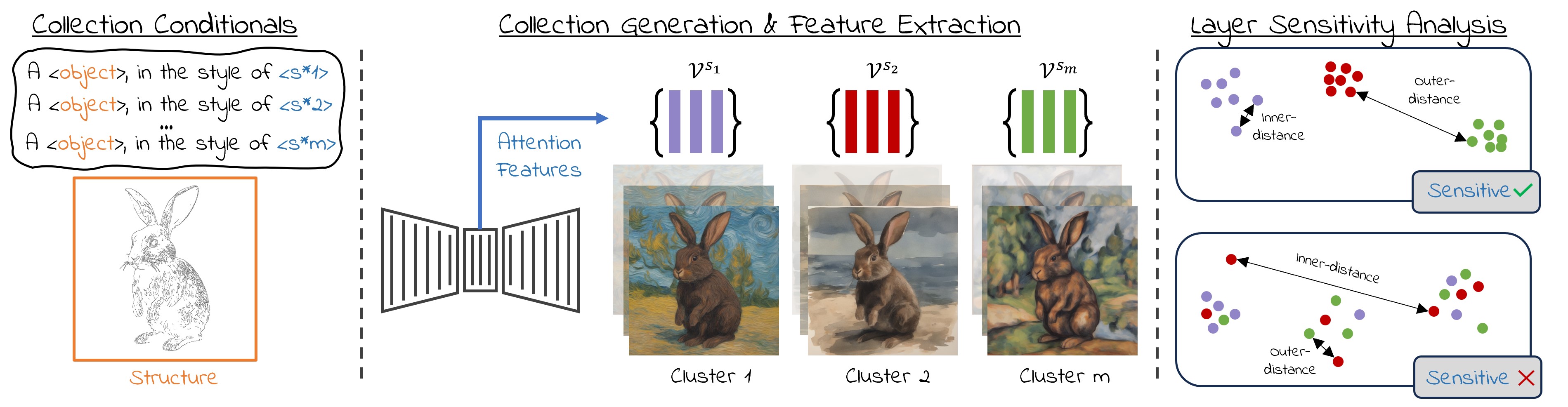}
    \caption{\textbf{Method Overview.} \textit{We generate a collection of images isolating a specific aspect of interest. Then, during generation, we extract self-attention features for each layer, at each timestep to represent the image. We calculate each layer's clustering score of all images sharing the same artistic aspect. The score measures both inner and outer cluster distances between all images in this representation. Finally, we rank each layer sensitivity to this aspect by its clustering score and use this information in our balancing strategy.}}
    \label{fig:method_minimal}
\end{figure*}

To investigate the relationship between content and style, we first choose 10 different artistic style conditionals (see supplemental file for details) and generate images with textual prompts either with or without edge conditional maps (using Canny edges of a real photograph). 
The textual prompts were divided into two categories: Easy and Complex. 
A prompt is considered complex based not only on its length and details but also on how far its content deviates from the typical subjects of the style.
For example, generating a `fantasy' theme is harder than a `rural-life' theme for an artistic style that follows Van Gogh, since his works often depict European rural life in the late $19^{th}$ century.
This division allows us to isolate the impact of textual conditional alone and better understand its limitations.

Next, we evaluate the style and content of all generated images. We follow recent works \cite{hertz2023StyleAligned, jeong2024visual, wang2024instantstyle} and evaluate \emph{style} using the cosine similarity of the Dino embeddings~\cite{caron2021} between the style image and the generated image. For \emph{content} we calculate the cosine similarity between Clip embeddings \cite{radford2021learningtransferablevisualmodels} of the text prompt and the generated image. 

StyleAligned method applies stylization across all SDXL self-attention layers, often resulting in images that are misaligned with the text prompt and exhibit noticeable artifacts. 
To alleviate this, we gradually reduce the number of layers for stylization randomly (from 70 to 0). In \cref{fig:perlimanery_fig} (left) we show both a plot of quantitative results of averaging across all generated images and a specific qualitative example.
As can be seen, reducing the number of layers significantly improves text alignment.
However, while text compliance improves, style fidelity decreases more or less linearly as fewer layers are stylized. 
Hence, we find that using all layers for stylization is an indication of \emph{Style Over-Conditioning} phenomenon.   

We also observe the opposite effect which we term \emph{Content Over-Conditioning}. This phenomenon is caused, for example, by using challenging (Complex) text prompts or by using conditional maps like edge and depth maps. In this case, even when using all layers for stylization there is reduction in style consistency, such as appears in the geometric aspects of the target style. In \cref{fig:perlimanery_fig} (right) we show both a plot of quantitative results of averaging across all generated images using an edge-map condition, and a specific qualitative example. This issue appears with or without image-based style conditioning (as illustrated in \cref{fig:geometry_test}), which suggests that the output is over-constrained by the content map rather than lacking more style conditioning.  

We used a similar analysis for B-LoRA method (see supplemental file) and found that the basic method often suffers from under-stylization, resulting in images that lack style. In addition, both methods suffer from a significant style loss when conditioned with challenging content inputs. 

These experiments suggest there is a content-style tradeoff that can possibly be alleviated by decreasing the strength of the different conditions for better balance between them.
This can be done by limiting the influence of style and content conditionals by applying them only to a sub-set of the layers.
However, simply selecting random layers does not provide stable results and generates inconsistent outputs. This motivated us to identify a smaller subset of style-sensitive layers that could achieve effective stylization while still maintaining good content alignment of various types (text, shape, geometry).  

\cref{fig:perlimanery_fig} shows our method's results of balancing content and style by judiciously selecting layers compared to the random layer selection. As can be seen, the potential in style alignment increases dramatically faster than that of the random selection and is significantly higher in all number of layers without sacrificing content information. 
Additionally, style levels are preserved when using an edge-map condition, unlike the random layer selection which drops dramatically compared to text-only conditioning.

In the following section, we expand our approach, which includes a novel method for analyzing style sensitive layers in diffusion models, and a way to apply this knowledge to balance the use of conditionals at inference time without the need for any additional training.

%% file: sec/4_method.tex
\section{Method}
\label{sec:method}

Hertz et al. \cite{hertz2023StyleAligned} demonstrate that style information is embedded in attention layers during generation and applies it for stylization. 
Building on this insight, we analyze attention layer data to assess style sensitivity at each timestep. Since this information is
encoded in the network, our approach implicitly extracts it using analysis of a collection of images created in varied artistic
styles to interpret these sensitivities (see \cref{fig:method_minimal}).


\paragraph{Image Collection.}
We first choose $m$ distinct styles such as Claude Monet and Winslow Homer, and for each style $s \, (1 \leq s \leq m)$ we create a cluster $C_s$ by generating $n$ images where a single stylistic aspect is repeated multiple times in different images 
by using various random seeds. This results in a collection of $m \times n$ images.

To isolate a particular artistic aspect within each cluster, we impose the condition that, apart from the aspect being examined, other aspects (such as the subject) must either remain constant across the collection or vary consistently throughout. 
This ensures that the analysis is focused solely on the chosen aspect, as will be further discussed \cref{sec:Analysis}.

\paragraph{Image mapping.}


Hertz et al.~\cite{hertz2023StyleAligned} align a series of generated images to a single reference style image by using Adaptive Instance Normalization (AdaIN)~\cite{huang2017adain} between the self-attention layers in the denoising UNet. 
For each layer, the Key and Query features are extracted from the style image and projected to an $[S, H, D]$ shaped tensor where $S$ is the number of pixels, $H$ is the number of attention heads and $D$ is the dimension of each head. They calculate the mean and std of $S$, resulting in two mean and std vectors of shape $H\times D$ which are used to normalize the features of the output images. 
We use the same method and calculate the vector of the means and standard deviations ($\mu, \sigma$) for each image in our collection from each layer before applying AdaIN. 
we treat this vector $\mathcal{V}$ of ($\mu, \sigma$) as the image representation in each layer for each timestep.



Our key idea for evaluating the sensitivity of each layer at each timestep is to measure how well the structure of the space created by mapping all images, using their representation $\mathcal{V}$, aligns with the real clusters in the collection of images. A Good clustering score means this layer (in this timestep) has high sensitivity to the aspect represented by the collection. We rank all layers based on the clustering score and use this ranking to select the top K layers for  this aspect conditional injection (see \cref{fig:method_minimal}).

\paragraph{Clustering Score.}
The representation of each image $i$ from style $s$, $\mathcal{V}_i^s$ is defined as a multi-dimensional vector of ($\mu, \sigma$). We treat this  representation as simple multi-dimension Gaussian distributions (where the covariance matrix has entries only on the main diagonal), and measure distances in feature space using Jensen-Shannon Divergence (JSD) metric, which is calculated as follows:
\begin{equation} 
JSD(\mathcal{V}_1,\mathcal{V}_2)= \frac{1}{2}(D_{KL}(\mathcal{V}_1,M) + D_{KL}(\mathcal{V}_2,M)) 
\end{equation}
Here, $D_{KL}$ is the KL-Divergence, $\mathcal{V}_1$ and $\mathcal{V}_2$ are the two Gaussian distributions being compared, and $M$ is the average Gaussian distribution defined by the means and standard deviations of $\mathcal{V}_1$ and $\mathcal{V}_2$.


Next, for a given layer $l$ at timestep $t$, we calculate the ``inner distance'' score by computing the average distance between all pairs of Gaussians within the same cluster $C_s$ (paintings with the same artistic style):
\begin{equation} 
D_{l,t}^{in} = \frac{1}{m} \sum_{s=1}^{m}\frac{1}{\binom{n}{2}}\sum_{i,j \in C_s}{JSD(\mathcal{V}_i^{s}, \mathcal{V}_j^{s})} 
\end{equation}
where $\mathcal{V}_i^{s}$ and $\mathcal{V}_j^{s}$ represent Gaussians for two different paintings within the same cluster $C_s$.

In addition, we calculate the ``outer distance,'' which is the average distance between Gaussians from different clusters (paintings with different artistic styles):
\begin{equation} 
D_{l,t}^{out} = \frac{1}{\binom{m}{2}n^2} \sum_{s_1 \neq s_2} {JSD(\mathcal{V}_i^{s_1}, \mathcal{V}_j^{s_2})} 
\end{equation}
where $\mathcal{V}_i^{s_1}$ and $\mathcal{V}_j^{s_2}$ are Gaussians from two different clusters $C_{s_1}$ and $C_{s_2}$.

\begin{figure*}[t]  
    \centering
    \includegraphics[width=\textwidth]{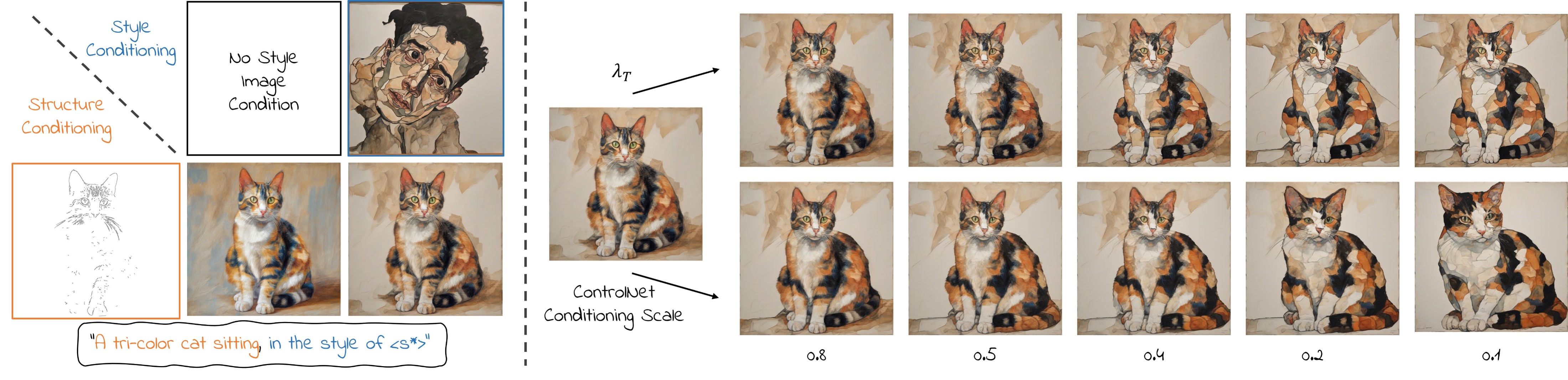}
    \caption{\textbf{Geometric Style Control}. \textit{In our experiments we found that geometric style is often lost when using structure conditional maps, whether using style conditionals or not (left). By limiting the structure conditional inputs only to non-sensitive geometric style layers we achieve a geometric style interpolation which combines geometric style gradually, using a timestep limiting parameter $\lambda_T$. Note that this is not possible to achieve by changing the default ControlNet scale parameter.
    Our method keeps the structure dictated by the structure conditional while gradually injecting geometric style to the image. (Please zoom in to view images better.)}}
    \label{fig:geometry_test}
\end{figure*}

Our objective is to identify layers that bring similar styles closer together and push different styles further apart. This means we aim to minimize $D_{l,t}^{in}$ and maximize $D_{l,t}^{out}$. Therefore, to evaluate the clustering, we calculate the ratio of inner to outer distances as follows (similar to Dunn index~\cite{dunnIndex74}):

\begin{equation} 
G_{l,t} = \frac{D_{l,t}^{in}}{D_{l,t}^{out}} = \frac{\binom{m}{2}n^2 \sum{JSD(\mathcal{V}_i^{s}, \mathcal{V}_j^{s})}} {\binom{n}{2}m \sum_{s_1 \neq s_2} {JSD(\mathcal{V}_i^{s_1}, \mathcal{V}_j^{s_2})}} 
\end{equation}

After computing $G_{l,t}$ for each layer $l$ and timestep $t$, we rank the layers from lowest to highest. 
This ranking indicates the sensitivity of each layer in each timestep to the artistic aspect being analyzed in the image collection. Then, to balance different conditions, we apply conditioning only on the $K$ most sensitive layers instead of all the layers. This allows for better balancing between multiple conditions and creates more stable results as we will demonstrate in \cref{sec:results}.

%% file: sec/5_Analysis.tex
\section{Analysis}
\label{sec:Analysis}

Using our method we conduct an analysis of SDXL layers attempting to identify layers which are sensitive to different conditioning aspects. We concentrate on two popular aspects: style and structure.
Each analysis was repeated 5 times using different objects to ensure robustness, and the final layer grading for each layer at each time step was averaged neglecting the best and worst grade to prevent outliers.

We conduct a similar analysis on the SD3.5-Large architecture~\cite{esser2024scalingrectifiedflowtransformers}. Details can be found in the supplemental file.

\subsection{Style Sensitivity}
\label{sec:Analysis-style}
To analyze style sensitivity we create a collection featuring 10 artistically varied styles but using a single object for content which is also constrained by a canny edge-image. For each style we generate 5 images using text-prompts of the format "\textit{$<$content prompt$>$, $<$style prompt$>$}" which creates a collection with 50 images. A sample of the collection can be seen in \cref{fig:method_minimal}, while the entire collection is presented in the supplemental file.

Using this data we apply our method and find that the dependency on timestep is small so we rank layers for any timestep for style sensitivity. \cref{fig:perlimanery_fig} shows results of applying style conditioning on decreasing subsets of layers from 70 to 0 based on this ranking.
As can be observed, compared to a random layer choice, our ranking yields subsets with stronger style consistency. This results in better stylized output generation, while preserving content alignment and lowering the conditional burden of the style image.
To better balance the different conditions we found that in most cases conditioning only on less than half of the layers (K=30) yields best results.

\begin{figure}[t]
    \centering
    \includegraphics[width=0.949\linewidth]{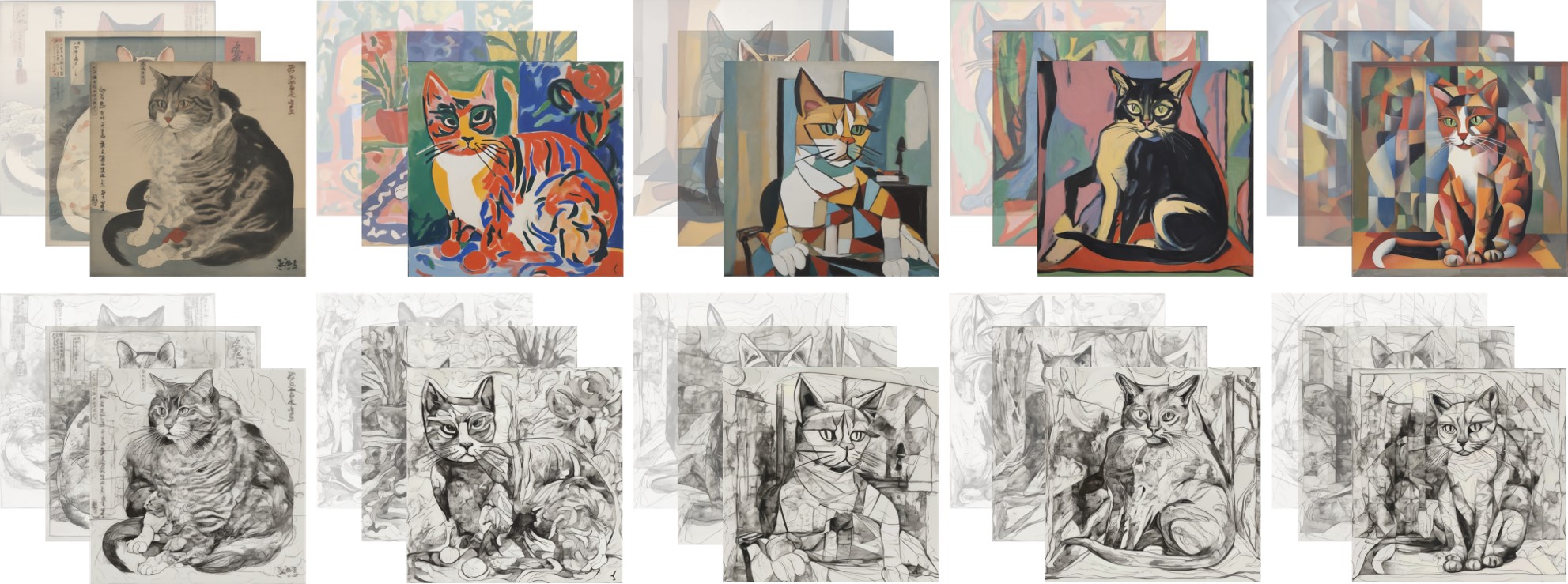}
    \caption{\textbf{Collection Example.} \textit{ Collection examples used for structure sensitivity analysis. The object depicted in the collection is constant (cat), where each cluster holds a different style for style conditioning (top). At the bottom, we use ``black and white ink'' style to normalize color balance for structure conditioning.}}
    \label{fig:collection_fig}
\end{figure}

\begin{figure*}[t]  
    \centering
    \includegraphics[width=\textwidth]{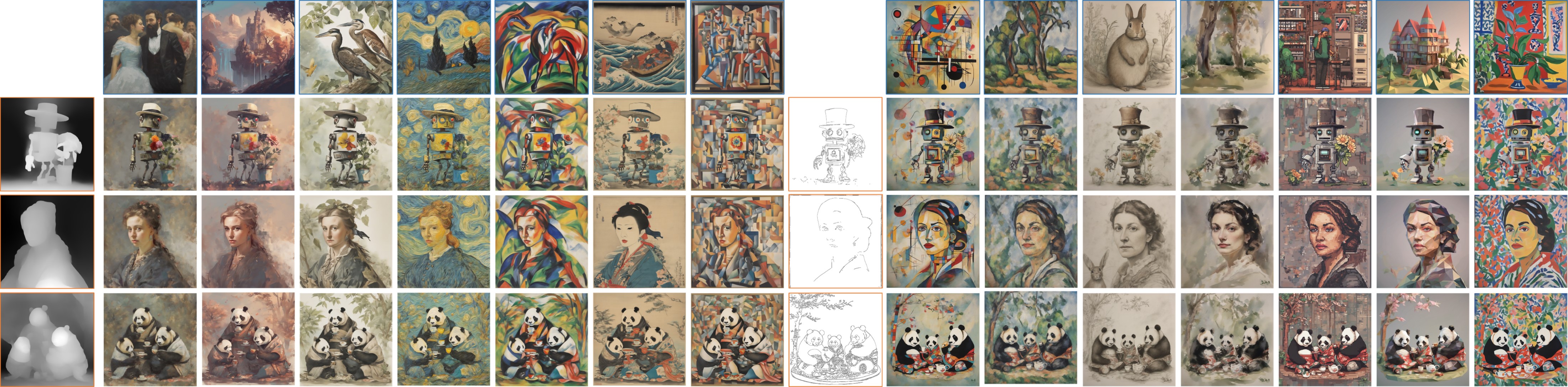}
    \caption{\textbf{Results.} \textit{Example results generated by balanced-StyleAligned over various styles (top). Prompts used from top to bottom: ``A robot wearing a fedora holding a flower,'' ``A portrait of a woman,'' and ``A family of panda bears wearing kimonos and sharing some tea.''}}
    \label{fig:results_new}
\end{figure*}

\begin{figure*}[t]  
    \centering
    \includegraphics[width=\textwidth]{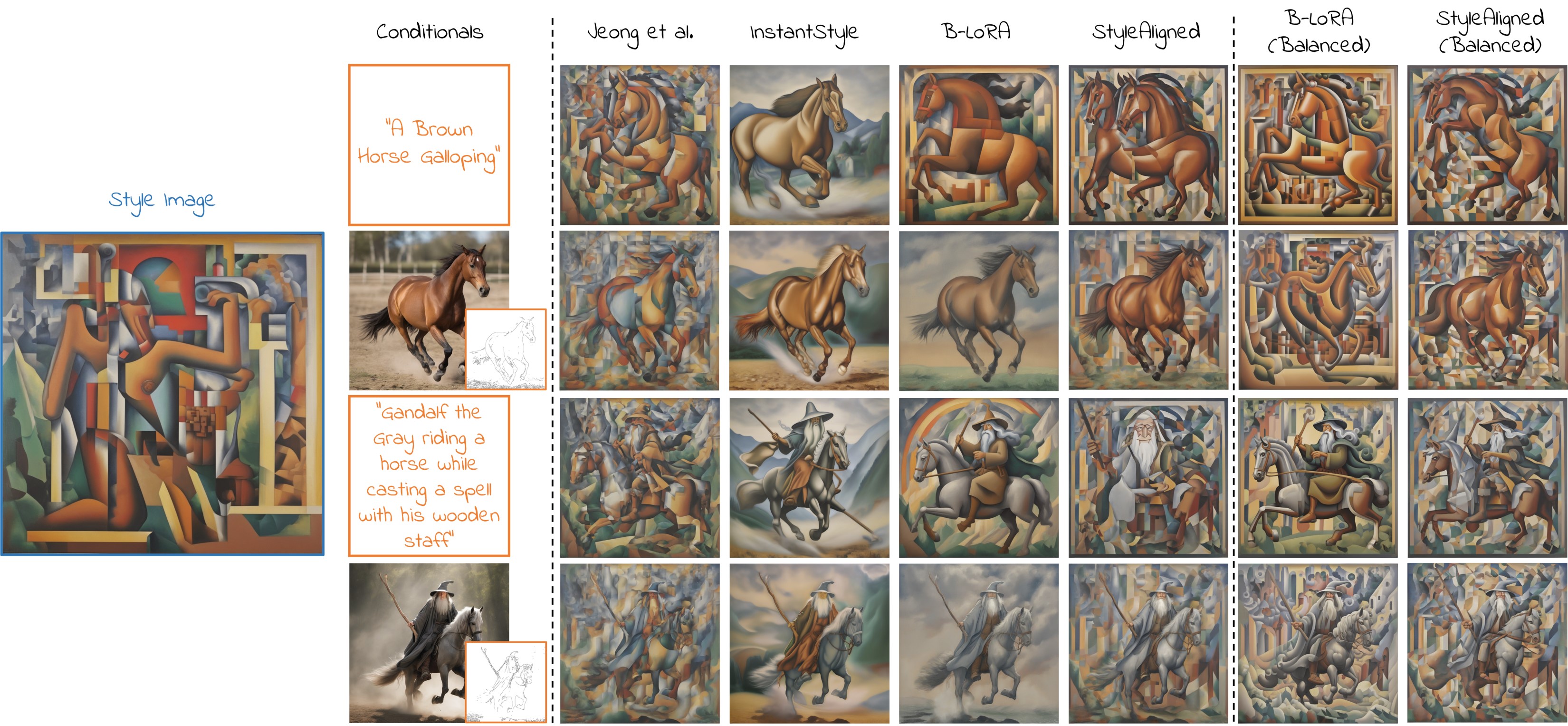}
    \caption{\textbf{Qualitative Comparison}. \textit{Comparing various conditional combinations: ``Easy'' vs ``Complex'' prompts (rows 1,2 vs. 3,4), Text only vs. Text and structure conditioning (rows 1,3 vs. 2,4). As can be seen, both balanced methods achieve consistency over all conditioning combinations while imbalanced methods show inconsistent generation quality and in some examples content and style issues.}}
    \label{fig:quanl_comp}
\end{figure*}

\subsection{Structure Sensitivity}

Conditioning structure in image generation is usually addressed by providing edge maps or similar maps with ControlNet~\cite{zhang2023adding}. Structure is also linked to aspects of geometric style of an artist's work, like contours and brush effects. Therefore, to analyze structure sensitivity we use Canny edge-maps conditioning. We select 10 artists with strong geometric styles and create five images for each artist, using a single object prompt for consistency. To further isolate geometric style from color, we apply an ``Ink Drawing'' style. A sample of the collection can be seen in \cref{fig:collection_fig}, while the entire collection is presented in the supplemental file. 

Applying our analysis method to this collection shows sensitivity in various Up layers at all timesteps. Reducing ControlNet input in these layers reveals geometric style changes as more timesteps are used. We conducted an ablation study (see supplemental file) to better understand these effects and discovered that the Up layers seem to handle the fine details from control maps, while Middle layers maintain overall structure alignment. Therefore, using our analysis we can rank the layers and timesteps according to structure sensitivity and apply conditioning (this time ControlNet input) only to a subset of the layers, allowing more balanced control with other conditions.
Note that this kind of geometric style control cannot be achieved by simply reducing the scale parameter of ControlNet (see \cref{fig:geometry_test}).

%% file: sec/6_Results.tex
\section{Results}
\label{sec:results}


Instead of choosing the number of layers K, we define two  interpolation parameters: $\lambda_S$ and $\lambda_T$, that control the percentage of layers used for style and structure respectively. This allows users to control different content-style balancing ratio, so we encourage tuning $\lambda_S$ and $\lambda_T$.
However, to be robust in our comparisons we set the layer subset size to 30 ($\lambda_S=0.43$) for both text and content image conditioned generations and limit the content control over 850 timesteps ($\lambda_T=0.15$) which induces geometric style and freedom. We find these values robust for various styles, including non-painting ones (see supplemental file.) We present results in \cref{fig:results_new} and in the supplemental file.



\subsection{Evaluation Details}
To evaluate our results we expand the evaluation set from \cref{sec:tradeoff}. For style conditionals, we use 32 different styles varying in origin (Europe, North America, South America, Asia), material (Oil, Watercolor, Digital, etc...) and style (Realism, Impressionism, Expressionism, Cubism, Anime, Pixel Art, etc...). For content conditionals, we use 10 ``Easy'' and ``Complex'' prompts each, and condition them with ``text only,'' Canny and Depth maps. For generating Depth maps we employ MiDaS~\cite{Ranftl2022} following~\cite{hertz2023StyleAligned}. We use 4 random seeds for generation of ``text only'' images, and two of these seeds for each image conditioning method, which uses 2 different content conditional images, one for each seed. Combining all this results in 5120 images for each method. We evaluate the style and content of each image according to the distance measures we defined in \cref{sec:tradeoff}, and average across all images for a given method. 



\subsection{Comparisons}
We apply our balancing strategy to StyleAligned and B-LoRA and compare our method to four style-based approaches: StyleAligned and B-LoRA (without balancing), InstantStyle, and Jeong et al.\ (we leave their pipeline analysis and balancing for future work) . We also compare to other methods\textcolor{blue}{~\cite{rout2024rbmodulation, cui2024instastyle}} in the supplemental file. As B-LoRA pre-trains residual LoRA weights prior to inference, we replace the trained layers to the ones analyzed by our method. For fairness, we base our decision on the same number of self-attention layers (10) used by B-LoRA, although we find that in some scenarios using a larger number of layers improves results. 

\begin{figure}[t]
    \centering
    \includegraphics[width=\linewidth]{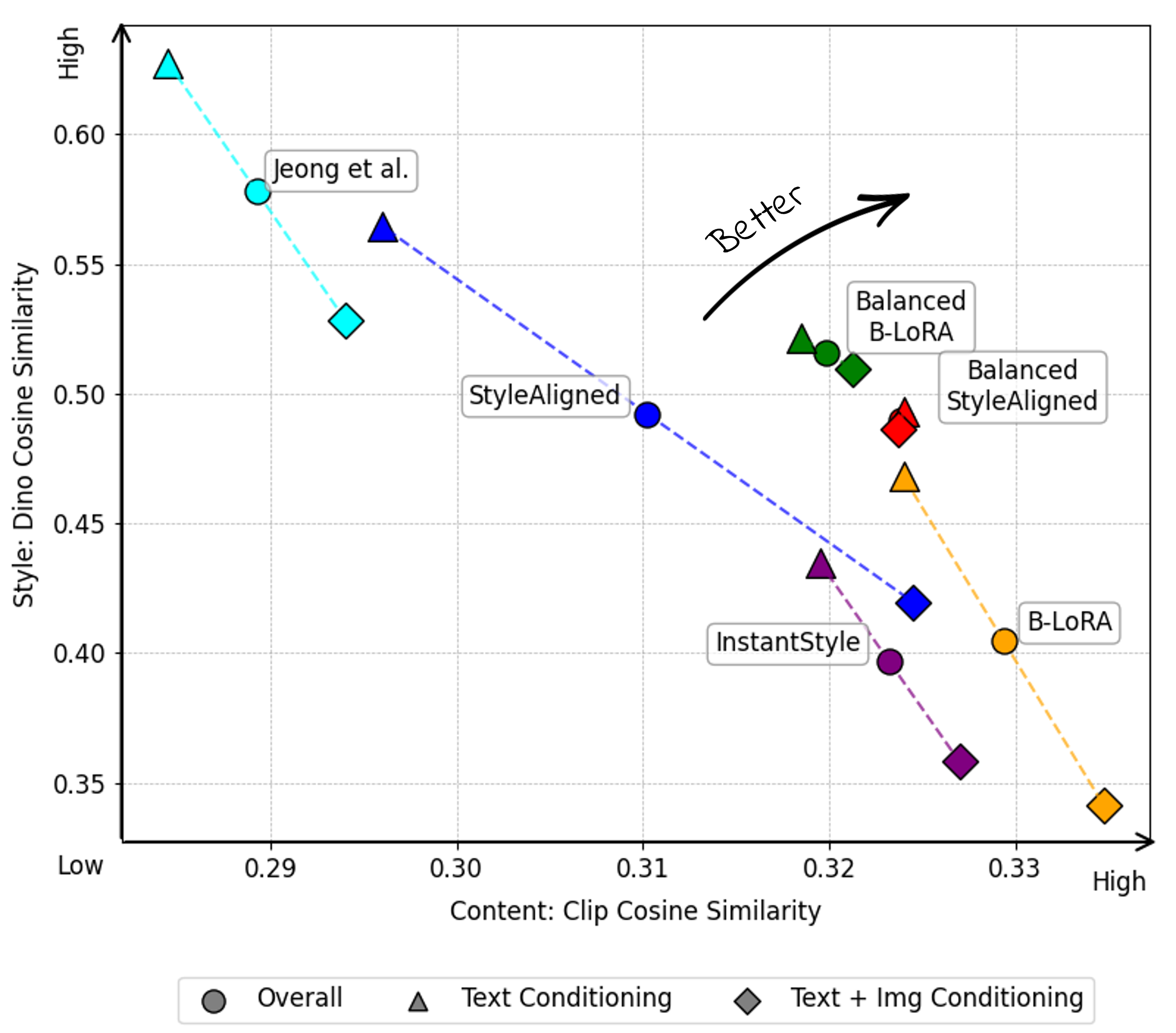}
    \caption{\textbf{Quantitative Comparison.} \textit{We compare recent methods to the balanced versions of StyleAligned and B-LoRA. The plot shows the content and style tradeoff improvement (middle circle) as well as the balancing effect over content image conditionals (triangle vs. rhombus).}}
    \label{fig:quant_comp}
\end{figure}

By definition, B-LoRA and InstantStyle are not intended to use a style prompt for generation, unlike the rest of the compared methods. For fairness, we evaluate B-LoRA and InstantStyle with style prompts like the other methods. B-LoRA shows a significant improvement in style with a slight decrease in content score. Since this minimally impacts its balance, we use it with style prompts for evaluation. However, for InstantStyle, the style improvement comes with a notable content reduction, significantly affecting balance, so we retain its original prompt for evaluation.

We report qualitative and quantitative results in \cref{fig:quanl_comp} and \cref{fig:quant_comp}, respectively. As demonstrated in \cref{fig:quanl_comp} our layer balancing strategy preserving both style and structure (right columns) and is more consistent over various conditioning combinations. Meanwhile, other methods shows inconsistent output quality, where some combinations show satisfying results while others show apparent issues such as general and geometric style loss, and content artifacts.

This balancing effect is also apparent quantitatively in \cref{fig:quant_comp}. While other methods show varying content and style quality on different conditioning combinations (triangle vs. rhombus), balanced methods show consistent high scores in both style and content similarity. We further evaluate these results using an additional style-representation using GRAM matrices~\cite{Gatys_2016_CVPR} (See supplemental file.)

\begin{table}[t]
    \centering
    \resizebox{\columnwidth}{!}{  
    \begin{tabular}{l|cc|ccc}
        \toprule
        \textbf{Observed (Expected)}  & Balanced & Imbalanced & Total votes & $\chi^2_{stat}$ & $p_{val}<$ \\
        \midrule
        \textbf{1. Multi-choice}      & 386 (210) &  244 (420) & 630 & 35.1 & $0.001$ \\
        \midrule
        \textbf{2. A/B B-LoRA}        & 195 (126)  &  57 (126) & 252 & 30.0 & $0.001$ \\
        \midrule
        \textbf{3. A/B StyleAligned}  & 185 (126)  &  67 (126) & 252 & 21.9 & $0.001$ \\
        \bottomrule
    \end{tabular}
    }
    \caption{\textbf{User Study.} \textit{Multi-choice 1/6 methods: 2 balanced vs. \cite{jeong2024visual, wang2024instantstyle, frenkel2024implicit, hertz2023StyleAligned} and two A/B tests: SA/B-LoRA vs. im/balanced.}}
    \label{tab:user_study}
\end{table}


\subsection{User Study}
We measure the balance effect on qualitative preference.
Hypothesis: people find balanced methods better than imbalanced counterparts due to content/style conditioning and aesthetics.
We evaluate our balanced methods, StyleAligned (SA) and B-LoRA, with three question setups:
\begin{enumerate}
    \item A multi-choice comparison to all other imbalanced methods~\cite{jeong2024visual, wang2024instantstyle, frenkel2024implicit, hertz2023StyleAligned} i.e., 6 options and 15 instances.
    \item A/B Test: B-LoRA vs. balanced B-LoRA, 6 instances.
    \item A/B Test: SA vs. balanced SA, 6 instances.
\end{enumerate}
Instances were sampled at random while keeping equal conditioning representation.
In both setups users were asked: \textit{``which of the images below follows both conditions better: (1) shows content described in `prompt', and (2) shows the style of `style image'.{''}} Examples can be found in the supplemental material.
The study spanned over 1,134 evaluations by 42 anonymous participants.
As detailed in \cref{tab:user_study}, all experiments concluded with a significant preference for the balanced versions in a $\chi^2$ test for independence.

\begin{figure}[t]
    \centering
    \includegraphics[width=\linewidth]{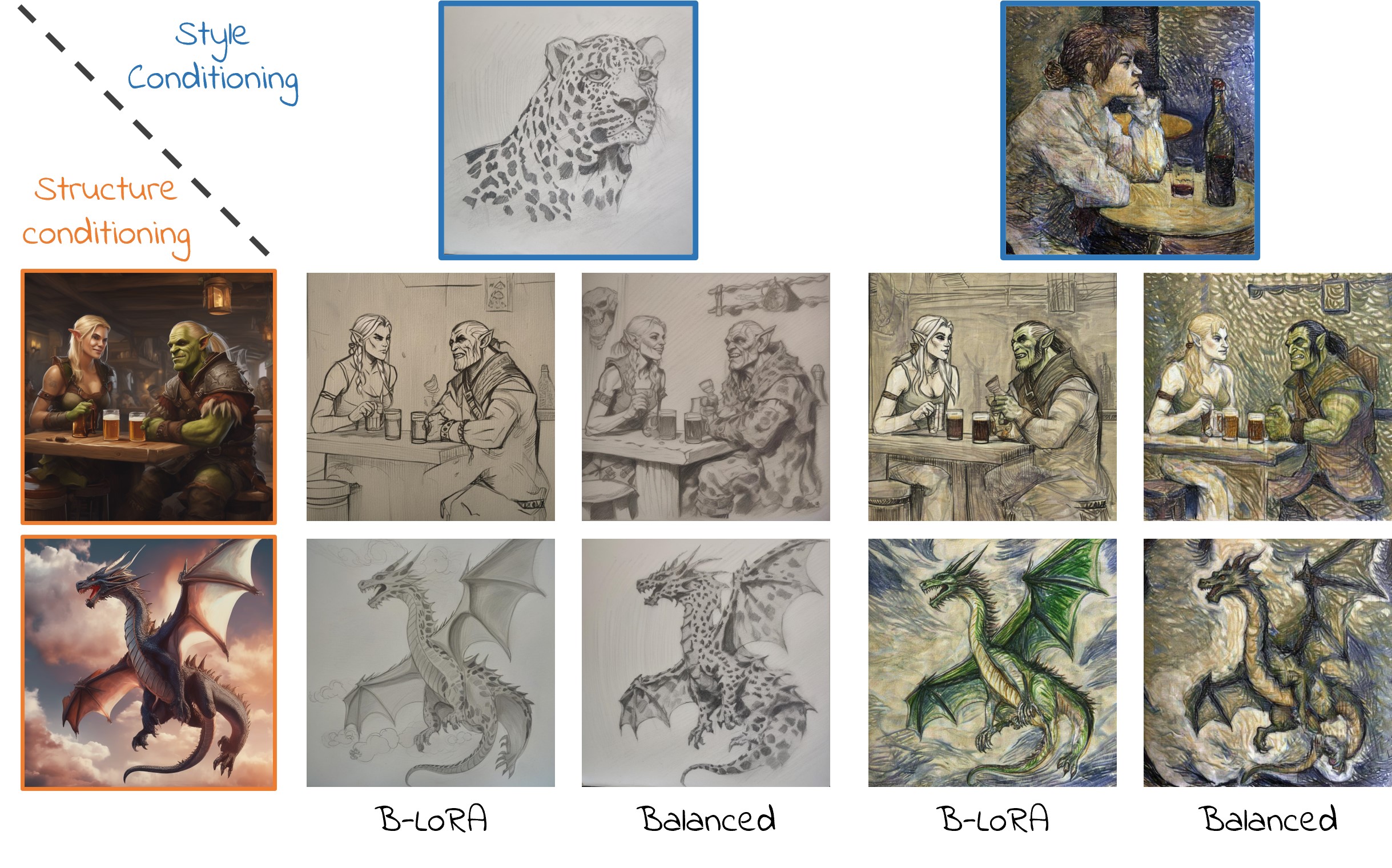}
    \caption{\textbf{Style Transfer.} \textit{A sample of style transfer using B-LoRA with an ungenerated style image (top row) and a content image (left column). Our balancing strategy improves style alignment and consistency for different content inputs. }}
    \label{fig:style_transfer}
\end{figure}


\subsection{Additional Applications}

We experiment with additional applications such as style transfer~\cite{Gatys_2016_CVPR, huang2017adain, Chen_2024_CVPR, johnson2016perceptuallossesrealtimestyle}, material-style generation~\cite{cheng2024zest}, and flexibly-conditioned content editing~\cite{liu2024smartcontrol, alaluf2023crossimage, bhat2023loosecontrol}.
We present style transfer results using balanced B-LoRA in Fig. \ref{fig:style_transfer}. Additional results and implementation details for all applications can be found in the supplemental file.

%% file: sec/7_Discussion.tex
\section{Discussion and Limitations}

In this paper we investigated the tradeoff between content and style conditionals and showed that over-conditioning prevents content and style alignment. We further present a novel method for analyzing sensitivity in attention layers of DDPMs by measuring the clustering score of the representation of a collection of images. We use this method to balance multi conditions by limiting conditioning on sensitive layers only, demonstrating improved results.

We found that the main limitations of our method arise from its dependence on the capabilities of the base model. For instance, when generating images with a style unfamiliar to the base model, the result may exhibit an unintended style, as the model lacks sufficient knowledge to properly generate the style reference, leading to a style mismatch. We demonstrate these limitations in the supplemental file.

\section*{Acknowledgment}
This work was partly supported by Joint NSFC-ISF Research Grant no. 3077/23


%% file: sec/X_suppl.tex
\clearpage
\setcounter{page}{1}
\maketitlesupplementary


\section{Appendix A. ---  Content-Style Tradeoff}
\label{sec:appendix_a_tradeoff}
In \cref{sec:tradeoff} we study the relationship between content and style and experiment with various conditioning settings. In this section we expand on our evaluation set and on additional experiments regarding B-LoRA~\cite{frenkel2024implicit}.

\begin{figure*}[t]  
    \includegraphics[width=\textwidth]{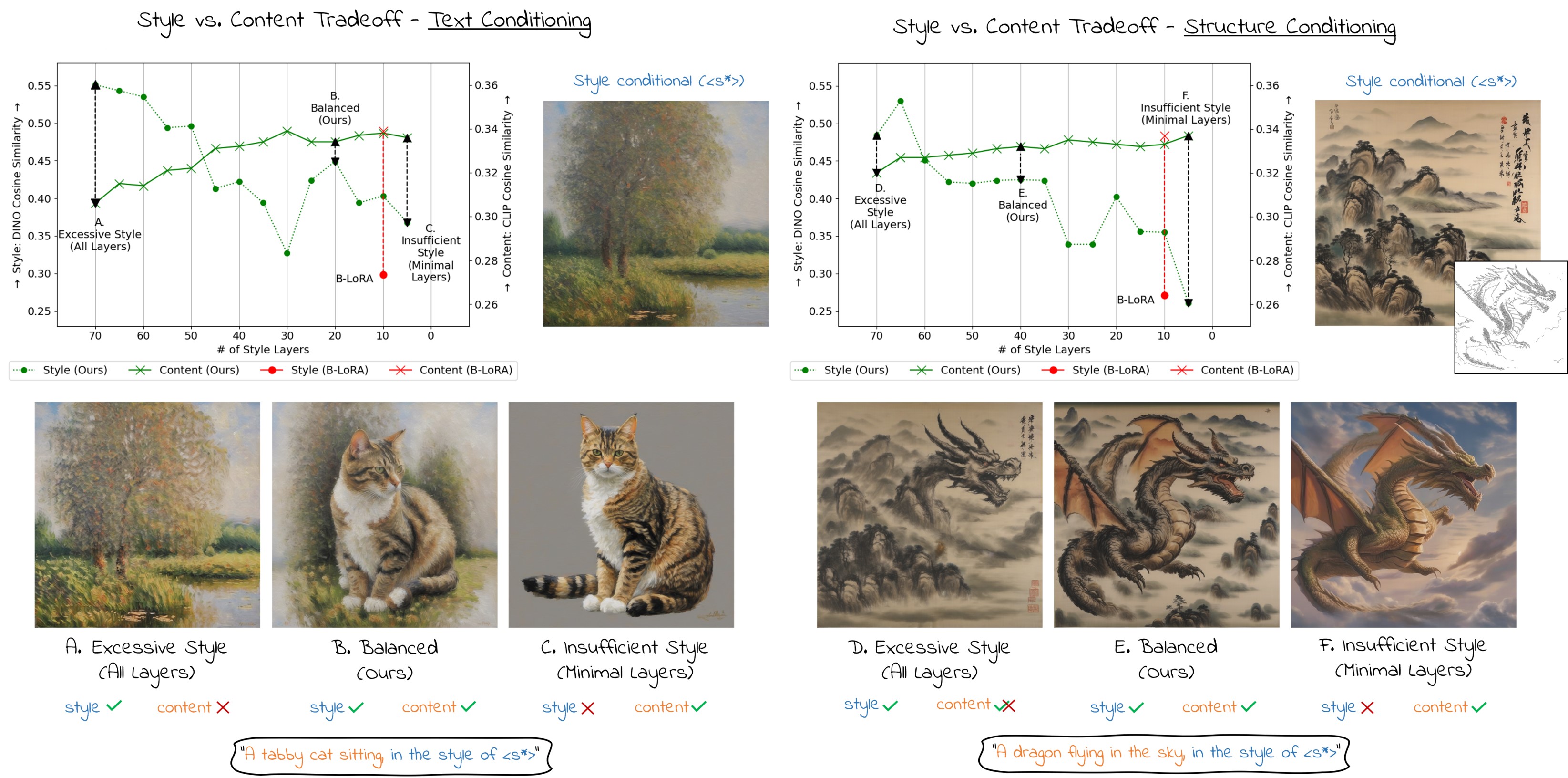}
    \caption{\textbf{Content-Style Tradeoff - B-LoRA.} \textit{Investigating the content-style tradeoff for B-LoRA vs. balancing using various number of stylization layers using our balancing strategy. As can be seen, using our strategy leads to balanced results for both `Text' and `Structure' conditioning and improves the overall generated image quality over imbalanced B-LoRA.}}
    \label{fig:perliminary_blora}
\end{figure*}

\subsection{Evaluation Set}
Our initial evaluation set contains 10 artistic styles: \textit{Beatrix Potter, Claude Monet, Egon Schiele, John Singer Sargent, Pablo Picasso, Studio-Ghibli, Utagawa Kuniyoshi, Vincent Van-Gogh, Winslow Homer,} and \textit{Xu Beihong}. For each style, we use five ``Easy'' and five ``Complex'' prompts for evaluation: \textit{``A bull moose,'' ``A grizzly bear,'' ``A dragon flying in the sky,'' ``A portrait of a woman,'' ``A tabby cat sitting,''} (Easy) and \textit{``A girl wearing a black and white striped shirt riding a bull moose in the Alaska wilderness,'' ``A family of panda bears wearing kimonos and sharing some tea,'' ``Two dragons, a green one and a red one, flying in a purple sky,'' ``A man wearing sunglasses and a woman watching the sunset from a mountain top,'' ``A ginger tabby cat riding a bicycle in Amsterdam next to a river''} (Complex).

In \cref{sec:results} we extend our evaluation set with 22 additional styles and 10 additional prompts. The additional styles are: \textit{Pixar, Pixel Art, Edvard Munch, Franz Marc, John James Audubon, Oswaldo Guayasamin, Henri Matisse, Wassily Kandinsky, Ilya Repin, Gustav Klimt, Voxel Art, Vector Art, Anime, Henri De Toulouse-Lautrec, Yoshitaka Amano, Cyberpunk, Concept Art, Low Poly, Gustav Courbet, Paul Cézanne, Jean Metzinger} and \textit{Georges Seurat}. The additional prompt are: \textit{``An old TV set,'' ``A colorful fishbowl,'' ``A house in a village,'' ``A bartender leaning on his bar,'' ``A brown horse galloping''} (Easy) and \textit{``A robot wearing a fedora holding a flower,'' ``A humpback whale floating in the sky carried by large colorful balloons,'' ``A fantasy castle with blue pointy rooftops located on a hill in a green valley,'' ``An orc and a blond wood-elf sitting in a tavern drinking beer as friends,'' ``Gandalf the Gray riding a horse while casting a spell with his wooden staff''} (Complex).

To generate the evaluation set we use 4 randomly chosen seeds: 10, 20, 9787, and 140592. For text-only generation we use all four seeds, for Canny conditioning we use 10 and 9787, and for Depth conditioning we use 20 and 140592.  

\subsection{B-LoRA Experiments}
In \cref{sec:tradeoff} we investigate the content-style tradeoff by using StyleAligned~\cite{hertz2023StyleAligned}. We expand this study for B-LoRA using the same evaluation set. Unlike StyleAligned, B-LoRA requires training residual weights prior to inference, which prevents applying stylization over a random set of layers for each evaluated image. Instead, we show the tradeoff between content and style using our balancing strategy and compare it to B-LoRA. Following \cref{sec:tradeoff} we use Dino~\cite{caron2021} and Clip~\cite{radford2021learningtransferablevisualmodels} embeddings to evaluate style and content, respectively, over various layer combination choices for both text conditioning and structure conditioning experiments.

We report both Qualitative and Quantitative evaluations in \cref{fig:perliminary_blora}. As illustrated, our strategy balances content and style for mutual conditioning. In the case of `Text Conditioning' (left) we can see that choosing style sensitive layers by our layer ranking yields a dramatic improvement in style over B-LoRA without sacrificing content, even when basing the stylization on only five self-attention layers. In this case we observe that choosing 20 layers yields a good balance between content and style. In the case of `Structure Conditioning' (right) using a structure control map yields more stability in content even for a high number of stylization layers. For this reason, we find that choosing 40 layers yields the optimal balance between content and style. 

In both cases we observe that using excessive style may lead to issues caused by content drift from the style image. When using a structure map (image D. in \cref{fig:perliminary_blora}) the impact can be marginal but when using a text condition alone (image A. in \cref{fig:perliminary_blora}) we can sometime lose the content of the image overall.

\begin{figure*}[t]
    \centering
    \begin{minipage}[b]{0.48\linewidth}
        \centering
        \includegraphics[width=\linewidth]{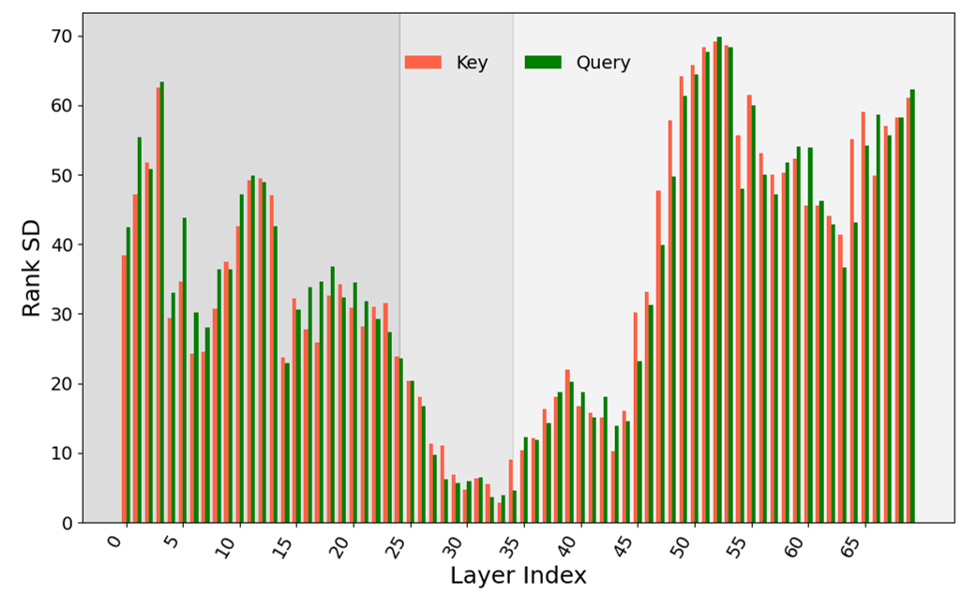}
        \subcaption{Style sensitivity - average rank over time}
    \end{minipage}
    \hfill
    \begin{minipage}[b]{0.48\linewidth}
        \centering
        \includegraphics[width=\linewidth]{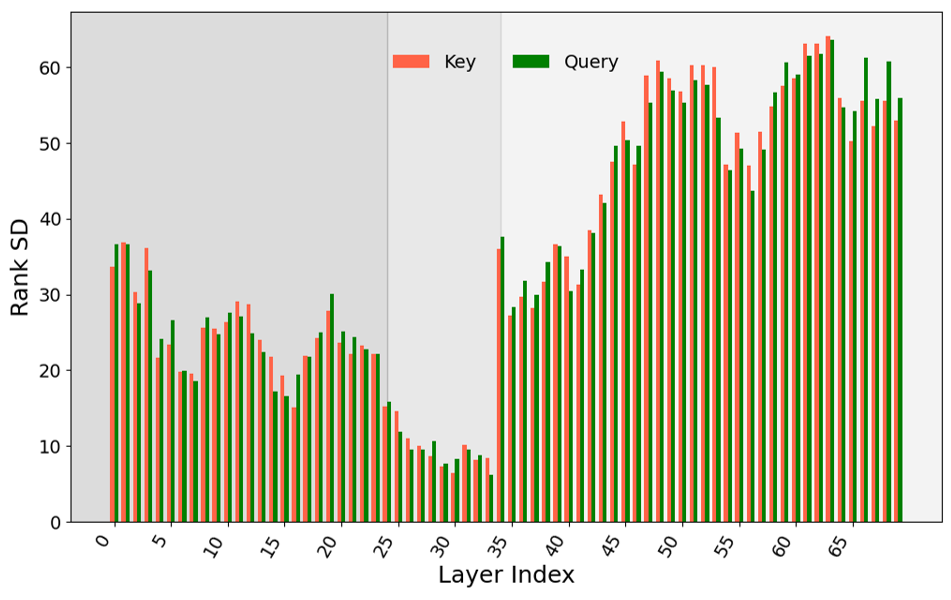}
        \subcaption{Geometric sensitivity - average rank over time}
    \end{minipage}
    
    \vspace{1em} 
    
    \begin{minipage}[b]{0.48\linewidth}
        \centering
        \includegraphics[width=\linewidth]{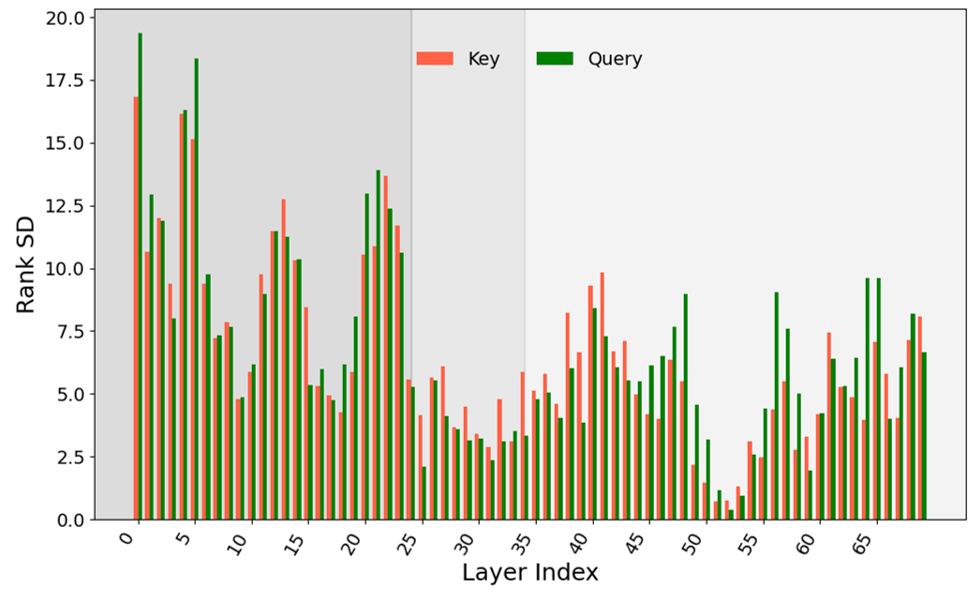}
        \subcaption{Style sensitivity - standard deviation over time}
    \end{minipage}
    \hfill
    \begin{minipage}[b]{0.48\linewidth}
        \centering
        \includegraphics[width=\linewidth]{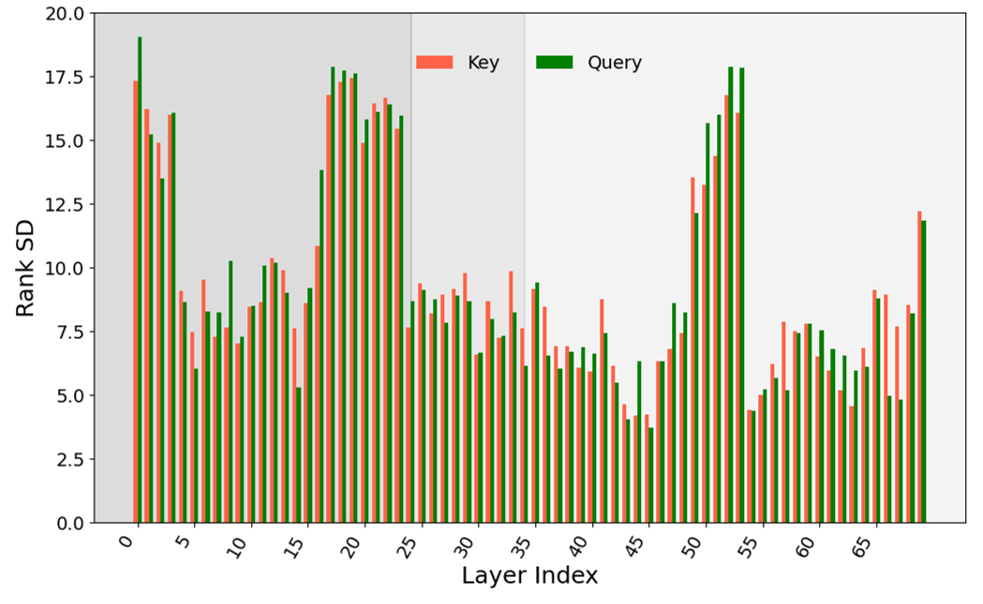}
        \subcaption{Geometric sensitivity - standard deviation over time}
    \end{minipage}

    \caption{\textbf{Average Layer Rank.} We show the average layer grade over all time steps for style and content sensitivity analysis. The top row depicts the first instance of each plot, and the bottom row duplicates them. As can be seen, various Up layers are important for both general style and geometric style. While geometric style seems to be more reliant on Up layers, some general style aspects seem to rely on Down layers. (Down, Middle and Up layers are divided by gray colored areas in the plot from left to right, respectively.)}
    \label{fig:rank_statistics}
\end{figure*}

\begin{figure*}[t]
    \centering
    \begin{minipage}[b]{0.48\linewidth}
        \centering
        \includegraphics[width=\linewidth]{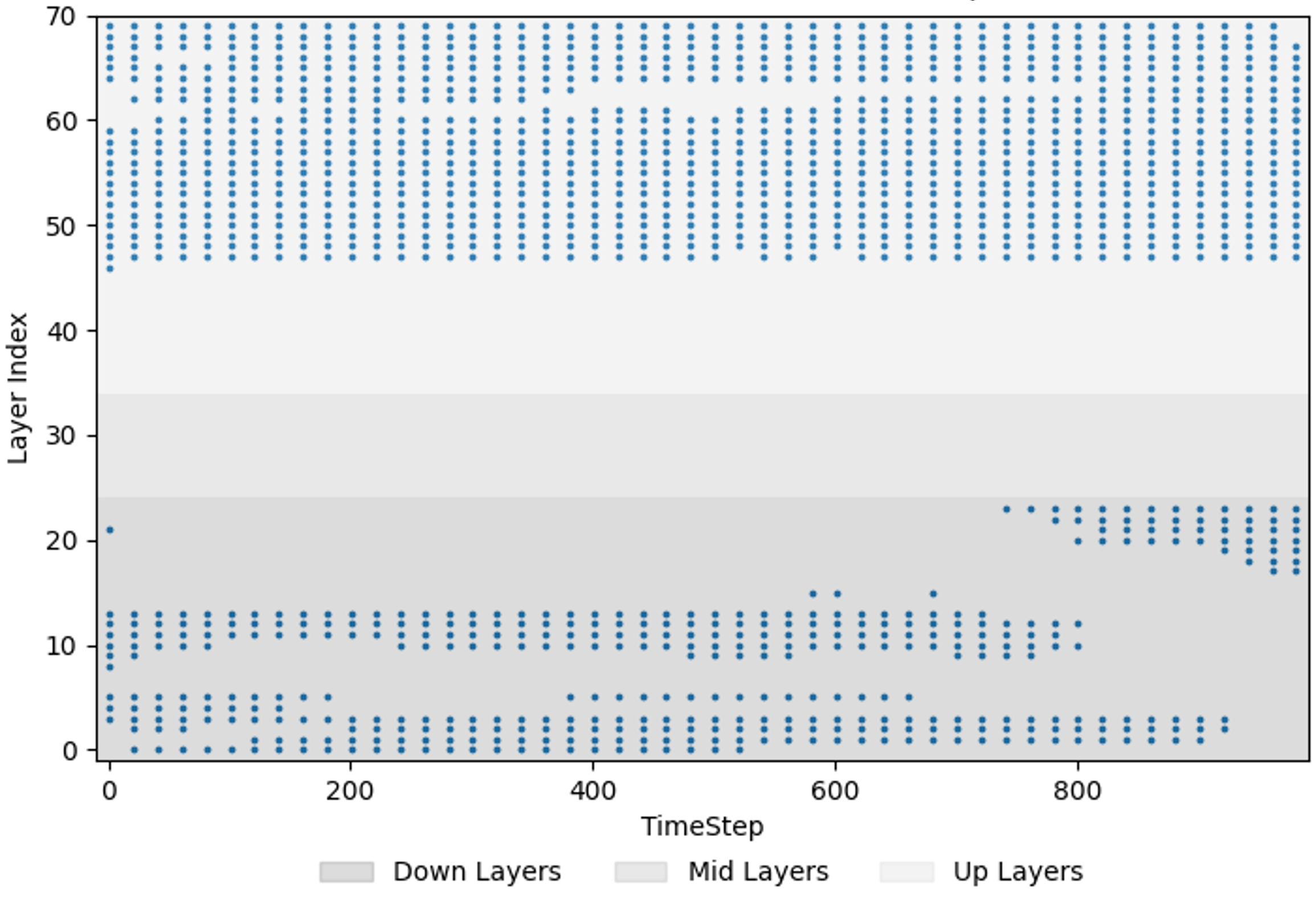}
        \subcaption{Style sensitive layer choice}
    \end{minipage}
    \hfill
    \begin{minipage}[b]{0.48\linewidth}
        \centering
        \includegraphics[width=\linewidth]{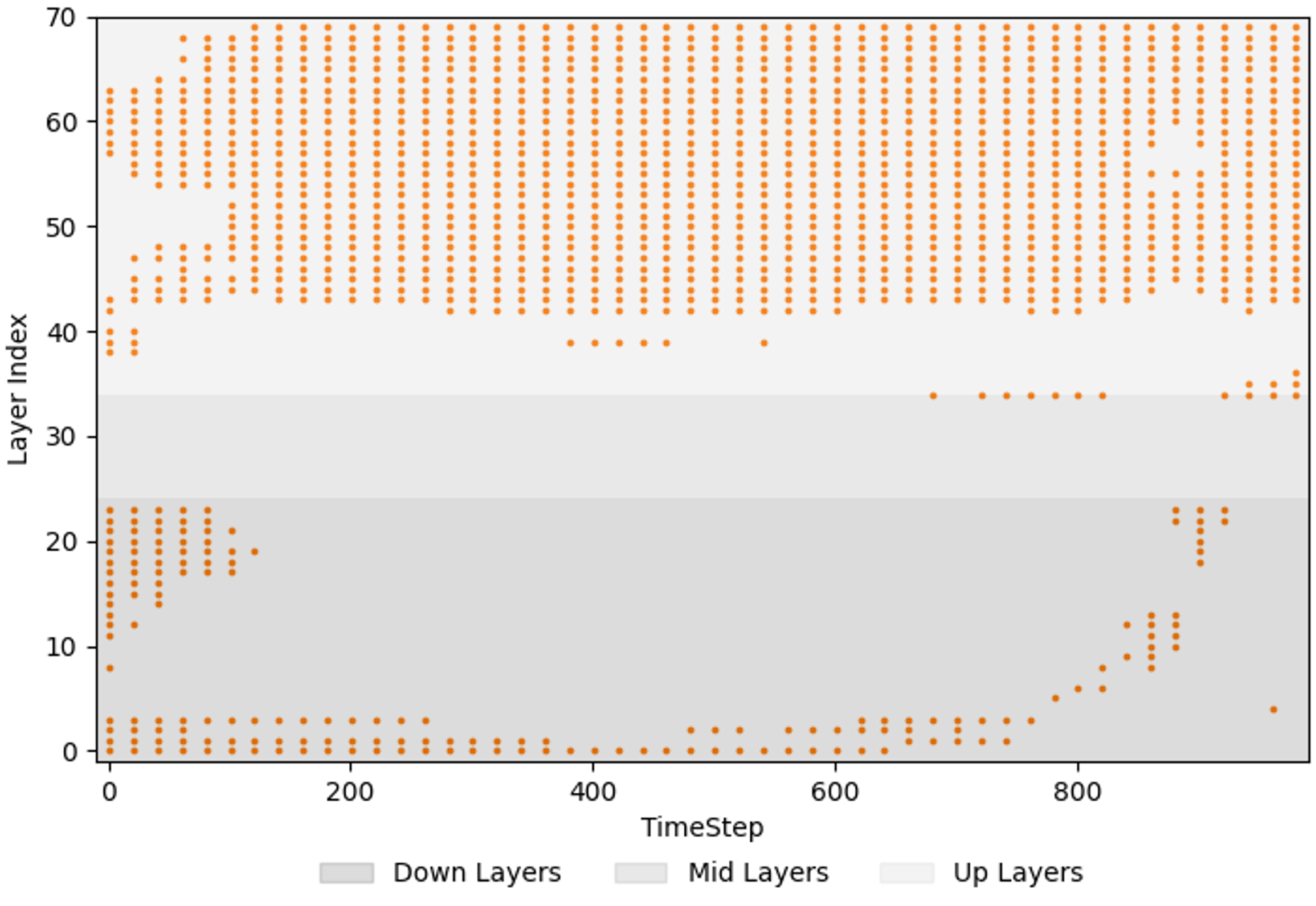}
        \subcaption{Geometry sensitive layer choice}
    \end{minipage}
    
    \vspace{1em} 
    

    \caption{\textbf{Layer Decision Example.} \textit{We show an example of the layer choice for $\lambda_S=0.43$ using 30 Key layers for style (left) and geometry (right). As can be observed the majority of layers show consistency over time while a some layers change for different timesteps.}}
    \label{fig:rank_choice}
\end{figure*}

\section{Appendix B. ---  Analysis}
\label{sec:appendix_b_analysis}

\subsection{Painting Collections}
\label{sec:supp_layer_rcollections}
Our style and content analysis is conducted over five collections each. For our style analysis we use five different objects and constrain their structure with a Canny map: \textit{Car, House, Rabbit, Bottle,} and \textit{Chair.} We generate 10 image clusters by various artistic styles: \textit{Vincent Van-Gogh, Claude Monet, Georges Seurat, Paul Signac, Edvard Munch, Winslow Homer, John Singer Sargent, Edward Hopper, Paul Cézanne,} and \textit{Berthe Morisot}. We choose these styles as they show variance in color and texture patterns but all have relatively realistic geometric style. The entire collections for \textit{Car} and \textit{Rabbit} are presented in \cref{fig:style_collection}.

The geometric sensitivity analysis was focused on limiting content conditionals from layers sensitive to geometric style. We choose five different objects: \textit{Cat, Wolf, Cow, Shark,} and \textit{Horse.} We choose to concentrate on animals as they tend to have more fluid interpretations in art paintings which is key for varying geometric style through the collections. We do not use a conditioning map to constrain structure as geometric style freedom is dependent on structure freedom. We generate 10 image clusters using the following styles: \textit{Jean-Michel Basquiat, Egon Schiele, Franz Marc, Vincent Van-Gogh, Ernst Ludwig Kirchner, Henri Matisse, Jean Metzinger, Edvard Munch, Pablo Picasso,} and \textit{Utagawa Kuniyoshi.} The entire collections of \textit{Cat} and \textit{Wolf} are presented in \cref{fig:cat_geo_col} and \cref{fig:wold_geo_col}, respectively, in both their color version and black and white version which was used in the analysis.

\subsection{Layer Rankings}
\label{sec:supp_layer_rankings}
Using our analysis method results with a ranking for each layer at each timestep. To better understand the ranking choices we show the mean and standard deviation of the layer rank over timesteps (\cref{fig:rank_statistics}). As can be seen, both style and geometry show a high correlation with the Up layers of the denoising UNet, while style seems to show a significant correlation also with Down layers.

We present an example of a choice of 30 layers of Key layers for both style and geometry in \cref{fig:rank_choice}. As can be seen the majority of layers show consistency over time while some layers change on various timesteps.

\begin{figure*}[t]  
    \includegraphics[width=\textwidth]{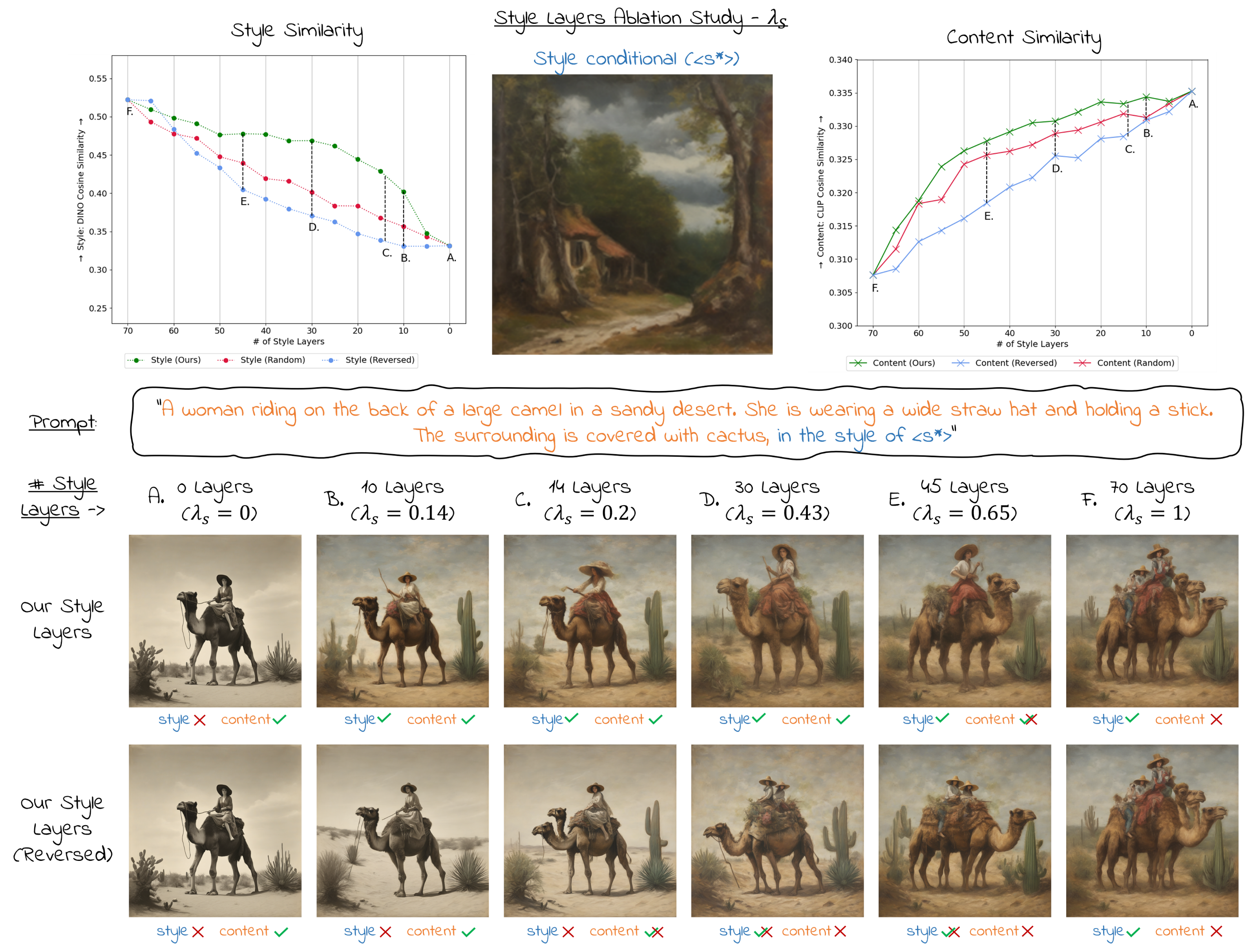}
    \caption{\textbf{Style Layers Ablation.} \textit{Ablation study for the ability of our sensitivity analysis method. We investigate and compare the method's ability to identify both style-sensitive and style-insensitive layers by using both layer rankings for stylization, where using insensitive layers is marked in \textcolor{blue}{blue} and using sensitive layers is marked in \textcolor{green}{green}. We show quantitative results (top) and compare both scenarios to a random layers selection (\textcolor{red}{red}) and show a visual example for demonstrative purposes (bottom).}}
    \label{fig:ablation_style}
\end{figure*}

\begin{figure*}[t]  
    \includegraphics[width=\textwidth]{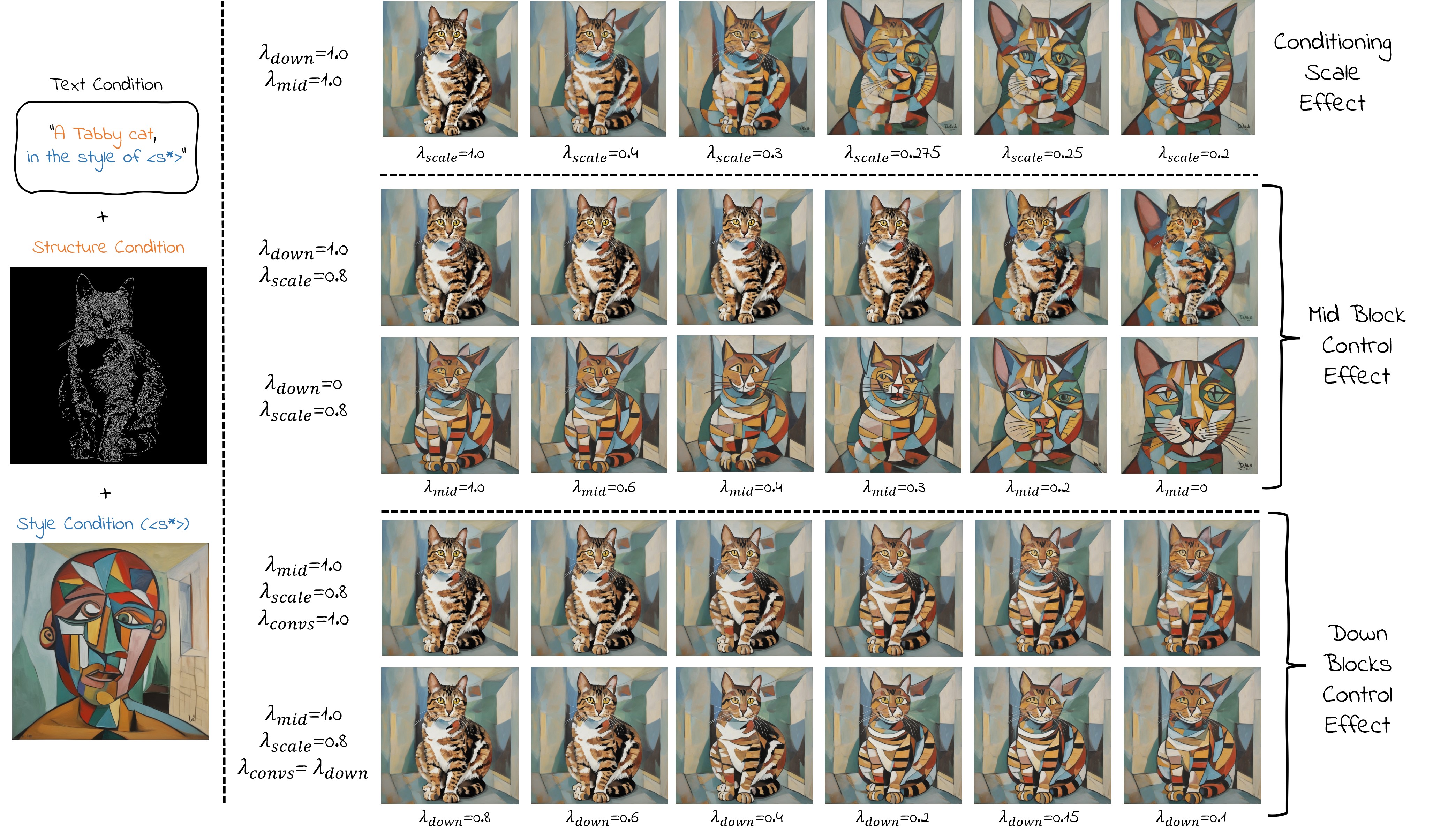}
    \caption{\textbf{Geometric Style Ablation.} \textit{Ablation study for our geometric scaling factor. We demonstrate the effect of $\lambda_{scale}, \lambda_{mid}, \lambda_{down}$ on the geometric style of the generated image. As can be seen, reducing results in a gradual decrease of fine details control and enables the model to gradually increase its geometric freedom which peaks around $\lambda_{down}$ value of 0.15. However, reducing control by lowering $\lambda_{scale}$ (top row) or $\lambda_{mid}$ (middle rows) results in abrupt loss of of general mask details around the value of 0.3, which leads to neglecting the content condition overall.}}
    \label{fig:ablation_geometry}
\end{figure*}

\subsection{Style Layers Ablation Study - $\lambda_S$}
\label{sec:supp_style_ablation}
In \cref{sec:tradeoff} of the main manuscript, we examine the trade-off between style and content in conditional image generation and provide an initial evaluation of the effectiveness of applying stylization to a subset of self-attention layers, as determined by our analysis method. In this subsection, we further explore our method's ability to identify style sensitivities by assessing the performance of layers marked as \textbf{not} style-sensitive. We compare their impact to both random selection and our previous results from \cref{sec:tradeoff}. To achieve this, we generate the evaluation set described in \cref{sec:tradeoff} using different subset sizes of style layers. However, instead of applying stylization to the $k$-most style-sensitive layers, we now apply it to the $k$-most \textbf{insensitive} layers.

Our experimental findings are illustrated in Fig. \ref{fig:ablation_style}. Similar to the approach in \cref{sec:tradeoff} of the main manuscript, we provide quantitative evaluations for both style similarity (top left) and content similarity (top right), alongside a qualitative example for visual demonstration.

As shown in the top left plot, applying stylization to layers identified as not style-sensitive (blue graph) results in a significant reduction in the style similarity of the generated images compared to stylization using the style-sensitive layers (green graph). Furthermore, applying stylization to these layers leads to lower style similarity than random layer selection, reinforcing the effectiveness of our method in correctly identifying both style-sensitive and style-insensitive layers. Notably, these insensitive layers appear to be entirely unrelated to style, as stylizing them negatively impacts the overall stylization quality. Additionally, using the insensitive layers causes a slower increase in style similarity, which only starts to improve around (E.), when the style-sensitive layers take effect. Observing the content similarity plot (top-right), we can confirm our hypothesis that injecting style information into style-insensitive layers (blue) is not only ineffective for style similarity but also degrades content similarity. This degradation results in lower content similarity compared to both style-sensitive layers (green) and randomly selected layers (red.)

The qualitative impact of these findings is visually demonstrated in the image sequence at the bottom of \cref{fig:ablation_style}. This sequence compares the interpolation effect of applying $\lambda_S$ on style-sensitive layers (top row) versus style-insensitive layers (bottom row). Each column ($A. - F.$), corresponding to points in the quantitative graphs, represents an increasing $\lambda_S$ value. Columns (A) and (F) show generated images for (A) no image style conditioning and (F) full layer style conditioning, both presenting a suboptimal result. The following observations can be made: (B) When using only 10 style-sensitive layers, the generated image already exhibits strong style representation while maintaining content integrity, whereas using style-insensitive layers results in no noticeable stylization effect. (C) Utilizing 14 sensitive layers maintains the previous quality, whereas using insensitive layers introduces content artifacts without achieving style alignment, likely due to injecting style information into content-related layers. (D) Applying 30 sensitive layers further improves both style and content alignment, while using insensitive layers results in some style improvements but also introduces content artifacts. (E) With 45 layers, stylization using our identified sensitive layers begins to introduce content artifacts, while using insensitive layers finally achieves reasonable style alignment as style and content-related layers start to blend in both cases.

\begin{figure*}[t]  
    \centering
    \includegraphics[width=\textwidth]{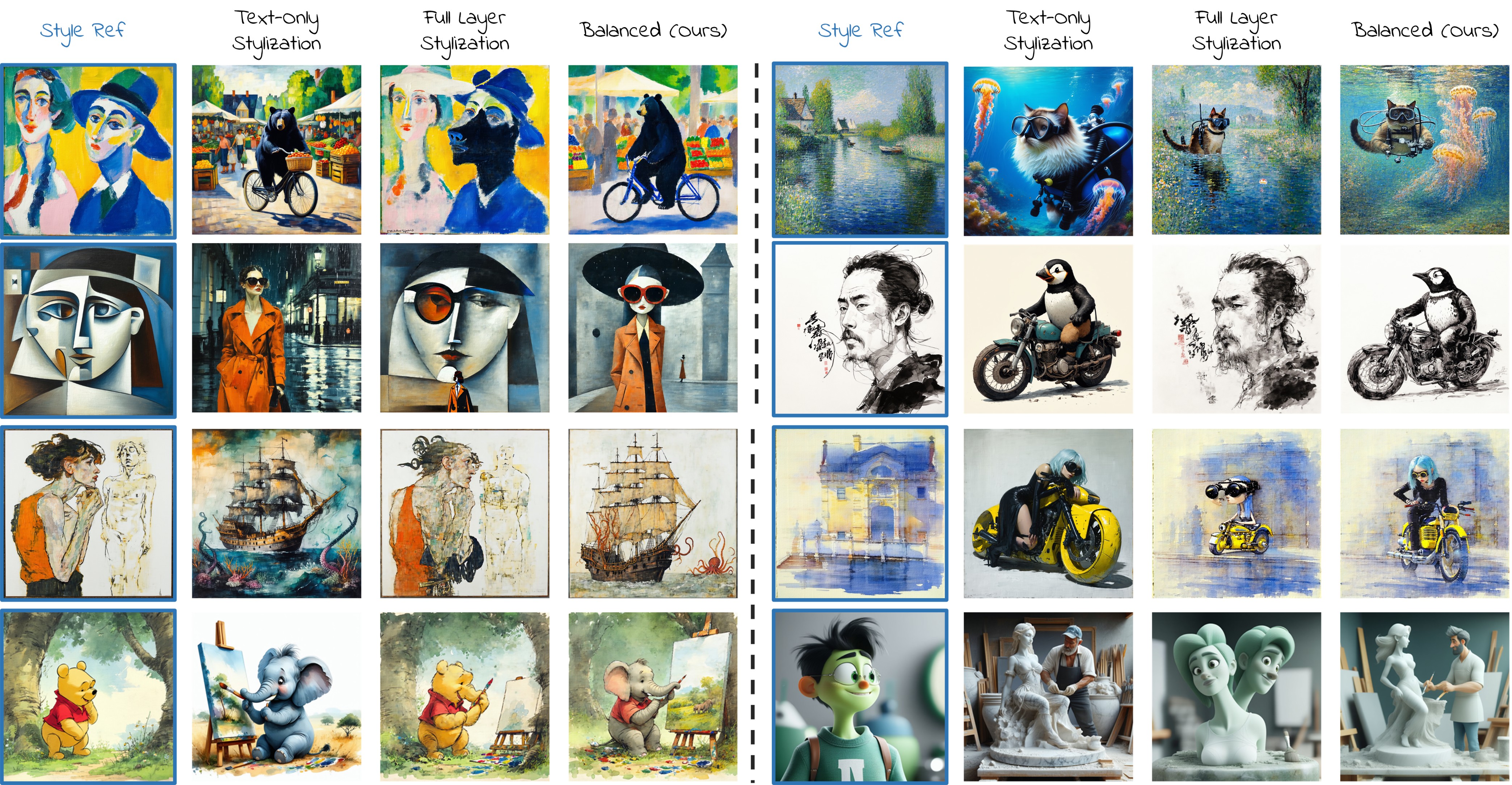}
    \caption{\textbf{SD3.5-Large Results.} \textit{Results for text-only stylization, full layer stylization and balanced conditioning with SD3.5-Large. Prompts: ``A black bear riding a bicycle in bustling market. The market is full of stands selling fruits and vegetables,'' ``a Siamese cat wearing scuba diving gear, horizontally scuba diving in a deep blue sea, watching a school of colorful jellyfish,'' ``A woman wearing a dark orange trench coat and large sunglasses walking in the cold streets of London,'' ``A penguin riding a motorcycle,'' ``A pirate ship sailing in the ocean and being attacked by the Kraken,'' ``A light-blue haired woman wearing a black attire and black steampunk goggles, leaning on a futuristic yellow motorcycle,'' ``An elephant painting the Savannah. He is sitting in front of a canvas holding a paint brush with his trunk,'' and ``A sculptor working in his studio. He is in the middle of sculpting a marble statue which starts to resemble a female figure.''}}
    \label{fig:results_sd35_supp}
\end{figure*}

\subsection{Geometric Style Ablation Study - $\lambda_T$}
\label{sec:supp_geometry ablation}
Following our analysis in \cref{sec:Analysis}, we conduct an ablation study to investigate the impact of the residual outputs of ControlNet on the generated images. ControlNet fine-tunes a copy of the denoising UNet encoder and extracts its outputs from the Down and Middle layers. These residuals are subsequently injected into the main UNet during generation at the Up and Middle layers, respectively. In our ablation study, as shown in \cref{fig:ablation_geometry}, we examine the effect of each of these layers and compare their influence to that of the default ControlNet conditioning scale, which reduces the conditioning effect on the output image. We define the parameters $\lambda_{scale}$, $\lambda_{down}$, and $\lambda_{mid}$ to control the default conditioning scale, the Down layer residuals, and the Mid layer residuals, respectively. The default parameter $\lambda_{scale}$ limits the conditioning effect by scaling the residuals, while $\lambda_{down}$ and $\lambda_{mid}$ restrict conditioning by applying it over fewer timesteps. Additionally, since some residuals are injected through convolution layers that are not analyzed by our method, we introduce $\lambda_{convs}$ to similarly limit the influence of convolutional-based layers.

As observed in \cref{fig:ablation_geometry}, each layer group exerts a distinct effect on the generation process. Adjusting $\lambda_{scale}$ and $\lambda_{mid}$ (top three rows) results in an uneven interpolation between full conditioning and no conditioning. In these cases, the generated images exhibit minimal changes across most $\lambda$ values (1.0 to approximately 0.3) before transitioning sharply (from~0.3~to~0.1) to images without any conditioning constraints. In contrast, interpolating over $\lambda_{down}$ (bottom two rows) reveals that the generated images progressively relax their adherence to the fine details of the conditioning structure image. This allows geometric style elements to emerge without compromising the broader structure of the image. Moreover, our experiments demonstrate that $\lambda_{convs}$ plays a significant role in incorporating geometric information in a visually pleasing manner.

These findings are consistent with our analysis in \cref{sec:Analysis}, which highlight the high sensitivity of the Up layers in the denoising UNet to geometric style. From this, we conclude that $\lambda_T$, which controls the conditioning injections in the Up layers over the timesteps of the generation process, enables interpolation over the amount of geometric style present in the output image.

\begin{figure}[t]
    \centering
    \includegraphics[width=\linewidth]{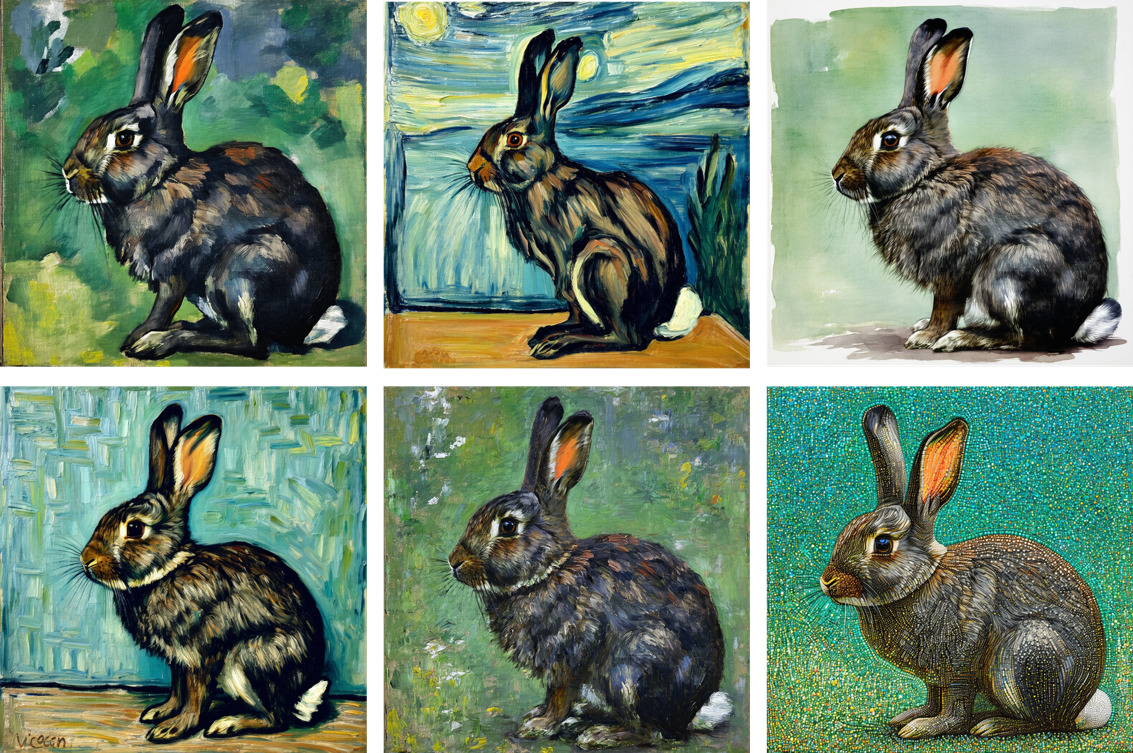}
    \caption{\textbf{Collection Examples.} \textit{Representatives from the style clusters generated during the SD3.5-Large analysis. Each sample represents a different style.}}
    \label{fig:sd35_clusters}
\end{figure}

\begin{table*}[t]  
\centering
\resizebox{\textwidth}{!}{  
\begin{tabular}{l|cc|cc|cc|cc|cc|cc|c|cc}
    \hline
    & \multicolumn{6}{c|}{Easy} & \multicolumn{6}{c|}{Complex} & \multicolumn{3}{c}{Easy + Complex} \\
    \hline
    & \multicolumn{2}{c|}{Text} & \multicolumn{2}{c|}{Depth} & \multicolumn{2}{c|}{Canny} 
    & \multicolumn{2}{c|}{Text} & \multicolumn{2}{c|}{Depth} & \multicolumn{2}{c|}{Canny} 
    & \multicolumn{2}{c}{Averaged} \\
    \hline
    Methods & Content & Style & Content & Style & Content & Style 
            & Content & Style & Content & Style & Content & Style 
            & Content & Style \\
    \hline
    
    Jeong et al.            & 0.277 & 0.625 & 0.278 & 0.563 & 0.293 & 0.485 & 0.292 & 0.630 & 0.289 & 0.579 & 0.316 & 0.487 & 0.289 & 0.578 \\
    InstantStyle            & 0.299 & 0.431 & 0.303 & 0.365 & 0.308 & 0.311 & 0.340 & 0.439 & 0.345 & 0.411 & 0.352 & 0.346 & 0.323 & 0.396 \\
    B-LoRA                  & 0.296 & 0.493 & 0.304 & 0.376 & 0.308 & 0.327 & 0.352 & 0.443 & 0.363 & 0.361 & 0.364 & 0.302 & 0.329 & 0.404 \\
    StyleAligned            & 0.273 & 0.592 & 0.295 & 0.459 & 0.300 & 0.408 & 0.319 & 0.537 & 0.348 & 0.429 & 0.355 & 0.383 & 0.310 & 0.492 \\
    \hline  
    
    B-LoRA (Balanced - 10 Layers)       & 0.291 & 0.548 & 0.291 & 0.533 & 0.293 & 0.528 & 0.346 & 0.495 & 0.349 & 0.479 & 0.352 & 0.499 & 0.319 & 0.515 \\
    B-LoRA (Balanced - 20 Layers)       & 0.286 & 0.575 & 0.288 & 0.547 & 0.292 & 0.540 & 0.339 & 0.522 & 0.344 & 0.509 & 0.350 & 0.511 & 0.315 & 0.537 \\
    StyleAligned (Balanced)             & 0.297 & 0.504 & 0.296 & 0.501 & 0.297 & 0.497 & 0.351 & 0.482 & 0.349 & 0.480 & 0.353 & 0.468 & 0.323 & 0.489 \\
    \hline  

\end{tabular}
}  
\caption{Comparison of methods across Easy and Complex prompts conditioned with and without Depth and Canny Conditioning.}
\label{tab:method_comparison}
\end{table*}

\section{Appendix C. ---  Stable Diffusion 3}
\label{sec:appendix_c_sd35}

\subsection{Style Conditioning}
Recently, the Stable Diffusion 3 (SD3) model family~\cite{esser2024scalingrectifiedflowtransformers} was released, offering new image generation diffusion models. Unlike SDXL~\cite{podell2023sdxlimprovinglatentdiffusion} which uses a UNet based on Self-Attention and Cross-Attention layers, the SD3 models are based on a Joint-Attention layers which processes both image information and text information. For this reason to apply balanced conditioning on these models we adapted the stylization ideas suggested by Hertz et al.~\cite{hertz2023StyleAligned}, which are Self-Attention based, to the Join-Attention architecture.

Like Hertz et al.~\cite{hertz2023StyleAligned}, we apply stylization by applying AdaIN~\cite{huang2017adain} between the attention features of a generated style image and the target image, and sharing the features of the Keys and Values on their projections. Since the Joint-Attention layer concatenates the Query, Key, and Value projections to the text encodings, we apply these operations before the concatenation to prevent changes in the text features. During our experiments we noticed that unlike Hertz et al. which applies AdaIN only on the Key and Query projections of the attention layers, applying AdaIN on the Value projections significantly contributes to the stylization of the target image and is key for achieving a satisfying result. For this reason we incorporate this change to our style conditioning algorithm for SD3.5-Large.

\begin{figure}[t]
    \centering
    \includegraphics[width=\linewidth]{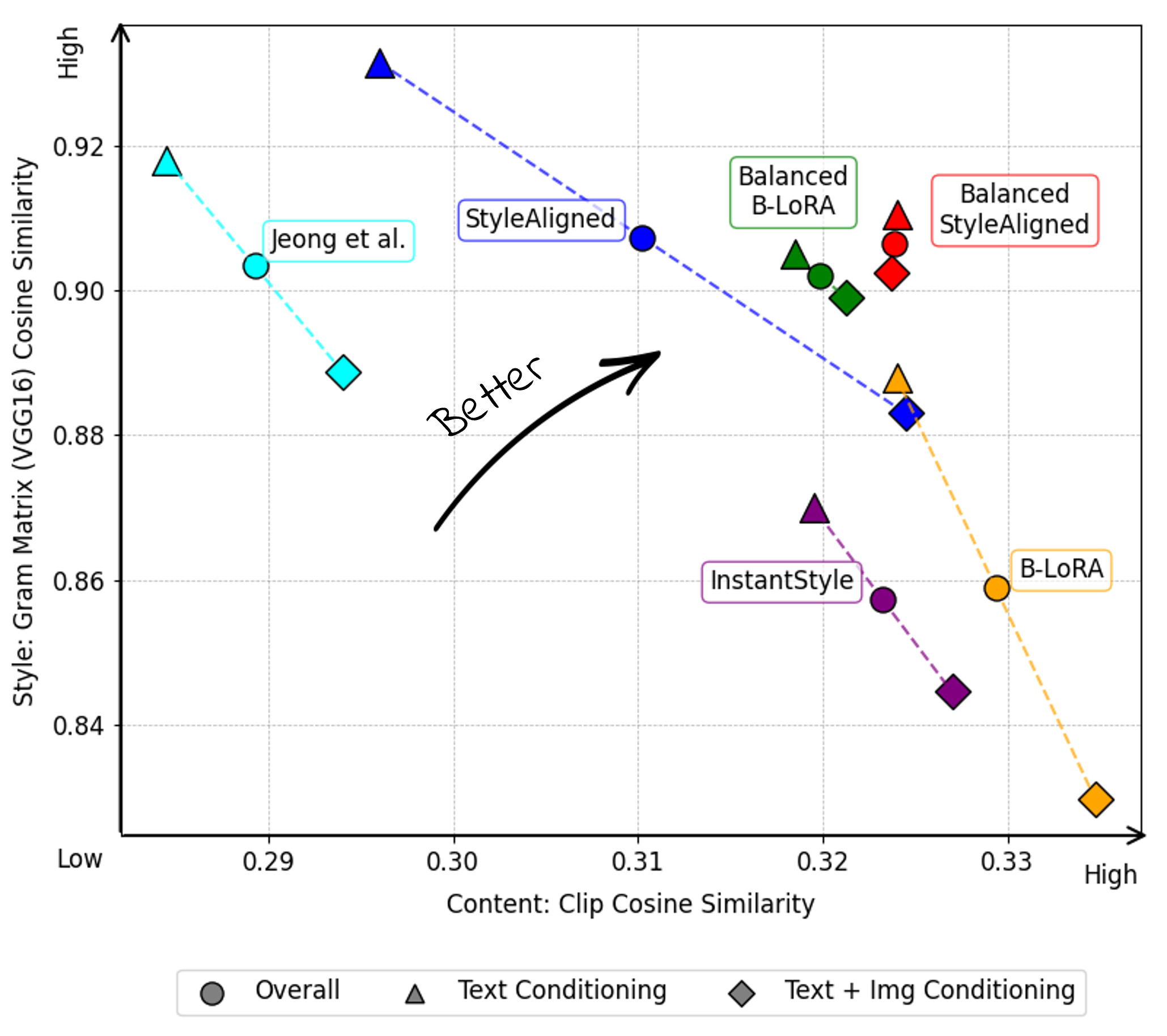}
    \caption{\textbf{Gram Based Evaluation.} \textit{Style and content evaluation using Gram-Matrix representation and Clip embeddings.}}
    \label{fig:gatys_quant}
\end{figure}

\subsection{Analysis}
We use our method presented in \cref{sec:method} to find SD3.5-Large style sensitivities as presented in \cref{sec:Analysis-style}. \cref{fig:sd35_clusters} shows examples from the collections generated during the analysis process where each example is taken from a different style cluster.

{As shown in and \cref{fig:results_sd35_supp}, SD3.5-Large, like SDXL, struggles with complex conditioning combinations. This is particularly evident in the "Text-Only Stylization" columns, where including only the artist's name in the prompt often results in style mismatches and, in extreme cases, the complete omission of the target style. Conditioning on all layers for style leads to significant content drift, whereas our balancing method effectively aligns the generated results with both the content and style of the reference image. The balanced results we present were achieved by applying style conditioning to 28 out of 38 joint-attention layers ($\lambda_S=0.73$), which we identified as the optimal setting.

\begin{figure}[t]
    \centering
    \includegraphics[width=\linewidth]{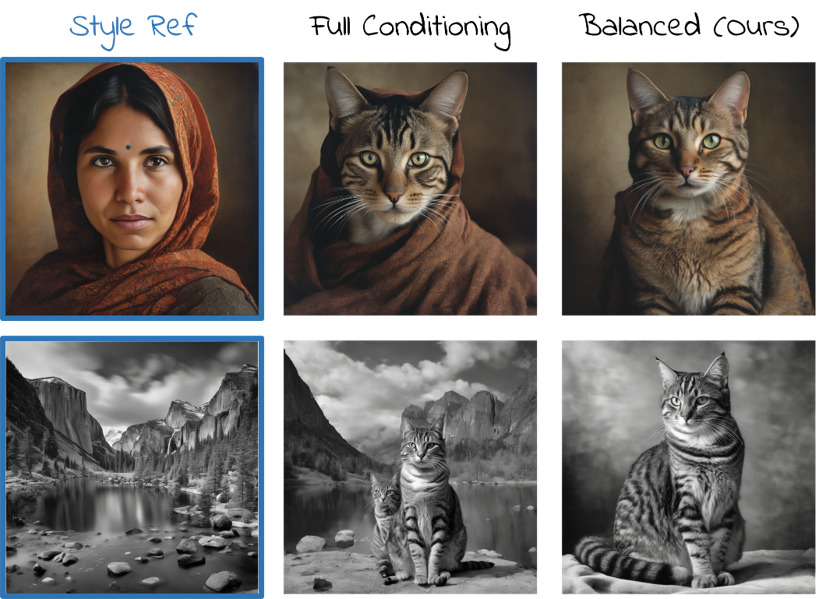}
    \caption{\textbf{Photographic-Editing Styles.} \textit{"A tabby cat". Style sensitive layers analyzed by our method are not limited to painting styles, but show sensitivities to other styles like photographic editing styles. Notice that using full conditioning may result with content drift and artifacts. }}
    \label{fig:photo_results}
\end{figure}

\section{Appendix D. ---  Results}
\label{sec:appendix_c_results}

\subsection{Qualitative Results}
To demonstrate the ability of our conditioning strategy we show additional results produced by the Balanced versions of StyleAligned and B-LoRA for ``Text Only'' (\cref{fig:results1}), ``Text + Canny'' (\cref{fig:results2}), and ``Text + Depth"' (\cref{fig:results3}).
We share additional qualitative comparisons between the balanced versions of StyleAligned and B-LoRA with the benchmark methods from \cref{sec:results} in \crefrange{fig:comparison_whale}{fig:comparison_dragon}. In addition, we compare ourself to two additional recent methods: RB-Modulation~\cite{rout2024rbmodulation} and InstaStyle~\cite{cui2024instastyle}. Since both methods provide access to their model only through an interactive web interface we could not evaluate their results using Canny or Depth conditioning. For this reason we refrained from using their methods in our main comparison in \cref{sec:results}. For fairness, we present a qualitative comparison in \cref{fig:additional_comp} using a style reference and a text prompt without any structure conditioning.

\subsection{Style Variation} To assess the generality of our method in identifying style-sensitive layers, we extend our experiments to styles beyond artistic paintings. \cref{fig:photo_results} illustrates the applicability of our style layers to photographic-editing styles. Notably, the same style layers identified for artistic paintings enable our method to generate photographs that adopt a reference style while preserving content integrity and avoiding unwanted artifacts and content drift from the style image.

\begin{figure}[t]
    \centering
    \includegraphics[width=\linewidth]{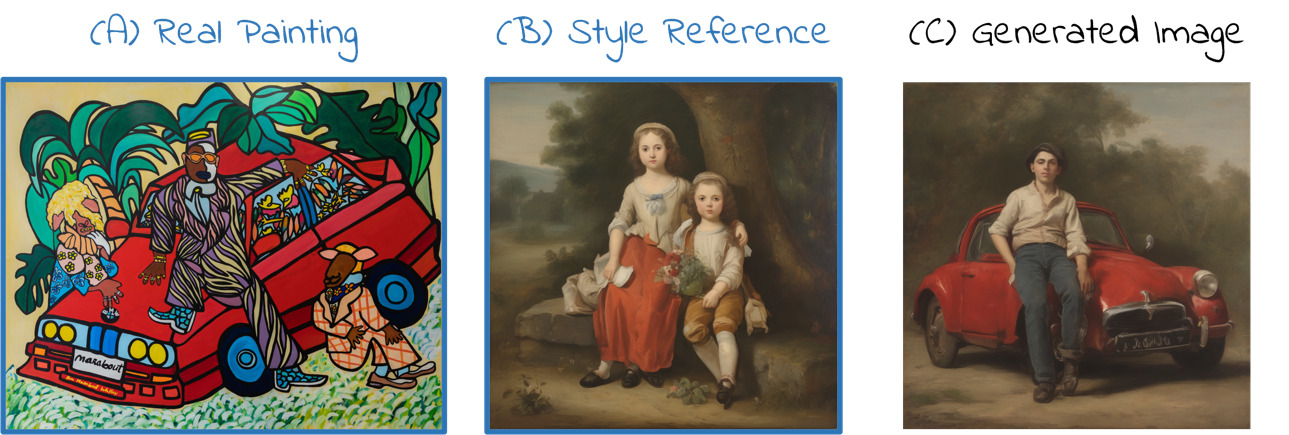}
    \caption{\textbf{Unfamiliar Styles.} \textit{A limitation example, where the style of an artist (A) is unknown to the base model, causing a mismatch in the reference image (B) which then results with a style mismatch in the target image (C), even when the content matches the artist. }}
    \label{fig:limitations}
\end{figure}

\subsection{Quantitative Results}
We present a breakdown of our quantitative results in \cref{tab:method_comparison}. We show the results over Easy and Complex prompts, for all conditioning types: ``Text Only,'' ``Text + Depth,'' and ``Text + Canny.'' We add an evaluation result for balancing B-LoRA based on 20 layers, which we found optimal for text conditioned generation. In addition we show the balanced version used in \cref{sec:results} which is based on 10 layers.

Since style representation is still an active area of research, we provide an additional evaluation of our results using Gram matrices~\cite{Gatys_2016_CVPR} as style descriptors instead of Dino~\cite{caron2021} features. As shown in \cref{fig:gatys_quant}, the Gram matrix based results are consistent with those presented in \cref{sec:results}, further highlighting the effectiveness of our method in achieving a balanced representation of both content and style.

\subsection{User Study Details}

\paragraph{Study Design and Participant Demographics}
The user study aimed to quantitatively evaluate the impact of balancing methods on the perceived quality of images conditioned on content and style prompts. A total of 42 anonymous participants took part, representing diverse backgrounds. The cohort included 62\% male participants, distributed across the following age groups: 16 participants aged 25--32, 10 aged 33--38, 10 aged 39--45, and 6 participants aged over 45. Professional affiliations spanned research and development (31\%), computer science graduate studies (24\%), professional artistry (19\%), and UX design (9\%), ensuring a broad spectrum of expertise relevant to the evaluation task.

\paragraph{Experimental Setup}
The study consisted of three tasks: a multi-choice comparison and two A/B tests, detailed in the main manuscript. Each task was designed to assess how well balanced and imbalanced methods align with both content and style, as perceived by users. The stimuli were generated by sampling from our dataset of text prompts and style reference images. Stratified sampling ensured a balanced representation of prompt complexity (``easy'' vs. ``complex'') and conditioning techniques (e.g., Canny, Depth, and Text-only). 

To eliminate biases, no style image was repeated across tasks, and the presentation order of images was randomized. Importantly, participants were not informed of the underlying generation method. The study was conducted online, with participants completing the evaluation independently, ensuring no researcher supervision or bias influenced the results.

\paragraph{Tasks and Protocols}
\begin{enumerate}
    \item \textbf{Multi-Choice Test}: Participants selected the best image from six options based on alignment with both the text prompt and style reference. Two of the six options in each instance were generated using balanced methods. The test encompassed 15 unique content-style pairings to ensure variety and robust statistical analysis. 

    \item \textbf{A/B Tests}: Each test involved binary comparisons between a method and its balanced counterpart. One test focused on B-LoRA, while the other evaluated StyleAligned. Both tests followed the same content-style alignment criterion and included six unique pairings for each method.
\end{enumerate}
Figures illustrating the test interfaces and sample questions can be found in \cref{fig:rb_user} and \cref{fig:ab_user}.

\begin{figure*}[t]  
    \centering
    \includegraphics[width=\textwidth]{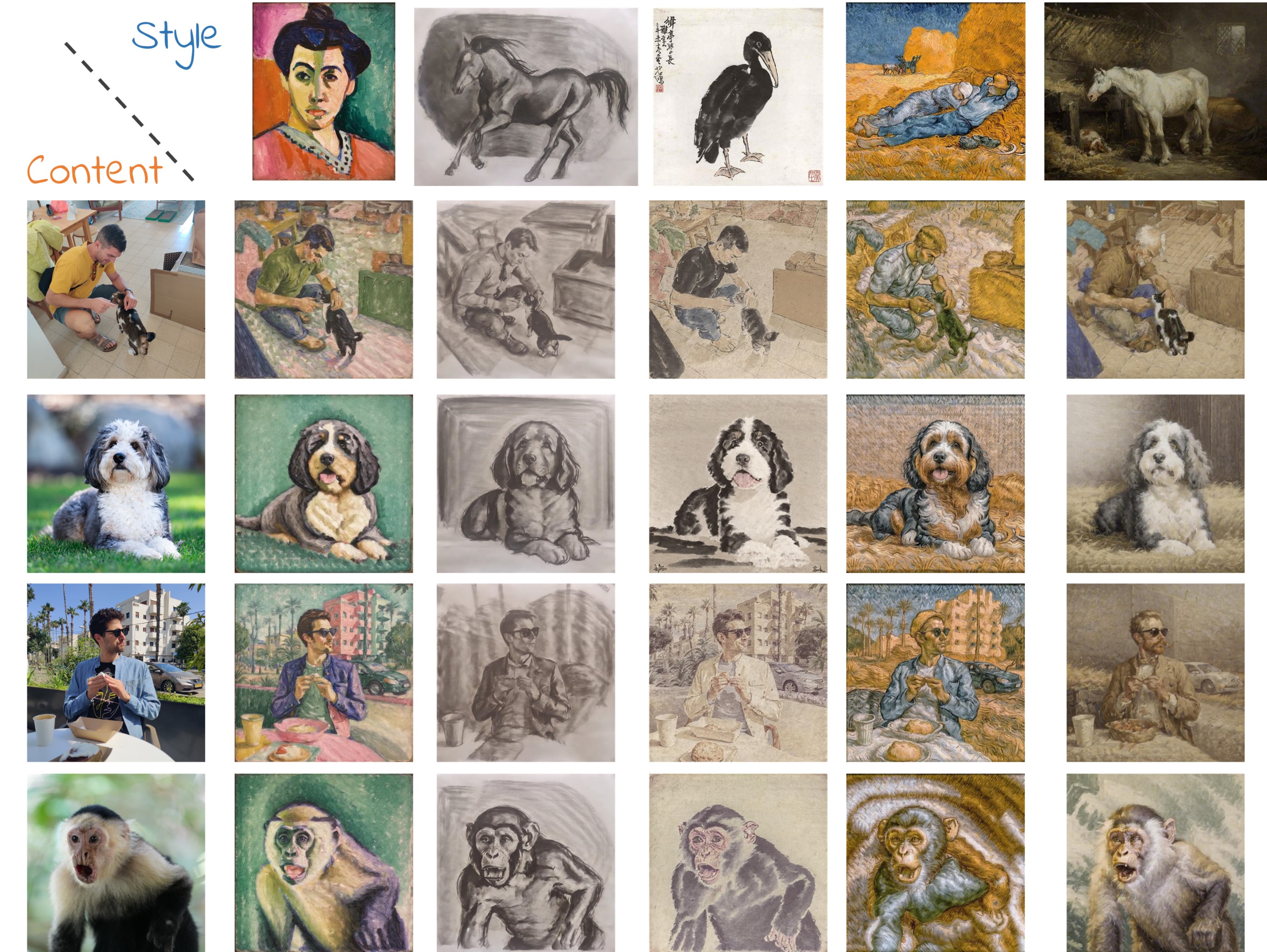}
    \caption{\textbf{Style Transfer.} \textit{Results generated using our balanced version of B-LoRA. Please zoom in for a better view.}}
    \label{fig:style transfer}
\end{figure*}

\paragraph{Statistical Analysis.}
Results were analyzed using a Chi-Squared Test for Independence to assess the preference for balanced versus imbalanced methods. The null hypothesis assumed no difference in user preference. For the multi-choice test, the expected probability of selecting balanced methods was set at $\frac{1}{3}$, based on their representation among the six options. The observed preferences significantly diverged from the null hypothesis, as shown in \cref{tab:user_study} of the main manuscript.

The study design and statistical robustness demonstrate a clear and significant preference for balanced methods, validating their efficacy in improving the visual alignment of content and style.

\subsection{Method Limitations}
As described in the main manuscript, the main limitations of our method arise from its dependence on the capabilities of the base model. For instance, when generating images with a style unfamiliar to the base model, the result may exhibit an unintended style, as the model lacks sufficient knowledge to properly generate the style reference, leading to a style mismatch. We demonstrate these limitations in \cref{fig:limitations}. As can be seen, the unique style of Enfant Précoce (A) is unknown to SDXL, thus the reference image (B) and output image (C) fail to match his unique style. This issue is not caused by a content misalignment issue, as the style fails to match the artist even when using a prompt that describes the content of an existing painting by the artist. In our demonstration, we use the prompt: \textit{``A man leaning on a red car surrounded by trees.''}

\section{Appendix E. ---  Additional Applications}
\label{sec:appendix_d_apps}

\subsection{Style Transfer}
To perform style transfer given two content and style images we use our balanced version of B-LoRA. For content alignment we use a Canny edge map as we find it the best option for preserving the structure and alignment of the given content image. For stylization we employ B-LoRA's approach and fine-tune residual LoRA weights on the given style image. Since B-LoRA does not change its stylization layer decision for each timestep, we rank the layers based on their average rank over all timesteps (see \cref{fig:rank_statistics}). In \cref{sec:results} we base our balanced version on 10 self-attention layers (20 including cross-attention layers) for fairness reasons, as it closely approximates the number of stylization layers used by B-LoRA. In practice we find that using a larger number of layers improves style fidelity. We experiment by using B-LoRA with various layer decisions (\cref{fig:perliminary_blora}), guided by our layer ranking and we find that basing our choice on the 20 best self-attention layers (40 with cross-attentions) strikes a fine balance between content and style. We show a quantitative ablation between B-LoRA and the two balanced variants in \cref{tab:method_comparison}. In addition, we show qualitative examples produced the balanced version of B-LoRA in \cref{fig:style transfer}. 

\subsection{Material Generation}
As shown in \cref{sec:results}, using our balancing strategy yields geometric style freedom when generating artistic images. Another result of this is better generation of material style. We show results in \cref{fig:material1} and \cref{fig:material2}. As can be seen, by applying our balancing strategy StyleAligned gains the ability to generate physical aspects of different materials even when conditioned on a content image. The regular version of StyleAligned forces unnecessary conditional information on the output on the content image, which results in patterns that do not match the material.

\subsection{ReStyle/ReContent}
Copying the works of old masters is a time-honored tradition in the art world, dating back to the origins of painting itself. This practice serves as a tool for artists to refine their techniques and develop their unique personal styles.
Throughout history, many renowned painters have engaged in this approach - examples include Vincent van Gogh, who copied works by Jean-François Millet, and Pablo Picasso, who reinterpreted works by Diego Velázquez such as Las Meninas.
This tradition has even given rise to several iconic artworks, such as Edgar Degas' studies of Old Masters like Nicolas Poussin and Rembrandt, or Francis Bacon's re-imaginings of Diego Velázquez's Portrait of Pope Innocent X.
Inspired by this classical method of artistic learning, we utilize our stylization approach, which enables the application of distinctive styles to the works of old masters - a process we call \textit{\textbf{ReStyle}}.
Additionally, our method extends the model's geometric flexibility, allowing for the re-imagining of an artwork's content in innovative ways - a feature we refer to as \textit{\textbf{ReContent}}.

To achieve this flexibly-conditioned image editing capability, we first use the original artwork as a structural condition, employing either a Canny or Depth map. We then generate the edited image using a relatively high style weight, $\lambda_S \approx 0.55$, a low content weight, $\lambda_T < 0.2$, and a descriptive text prompt. Setting these values for $\lambda_S$ and $\lambda_T$ allows for a strong resemblance to the artistic style of the style condition while providing geometric flexibility. This not only ensures a fine resemblance to the style condition but also enables content modifications to the original image through the text prompt. Examples are shown in Figs. \ref{fig:restyle_recontent1} and \ref{fig:restyle_recontent2}. As demonstrated, our approach effectively edits both the style and content of the original image while preserving its underlying structure and general characteristics, even when the generated content shift is significant.


\begin{figure*}[t]  
    \centering
    \includegraphics[width=\textwidth]{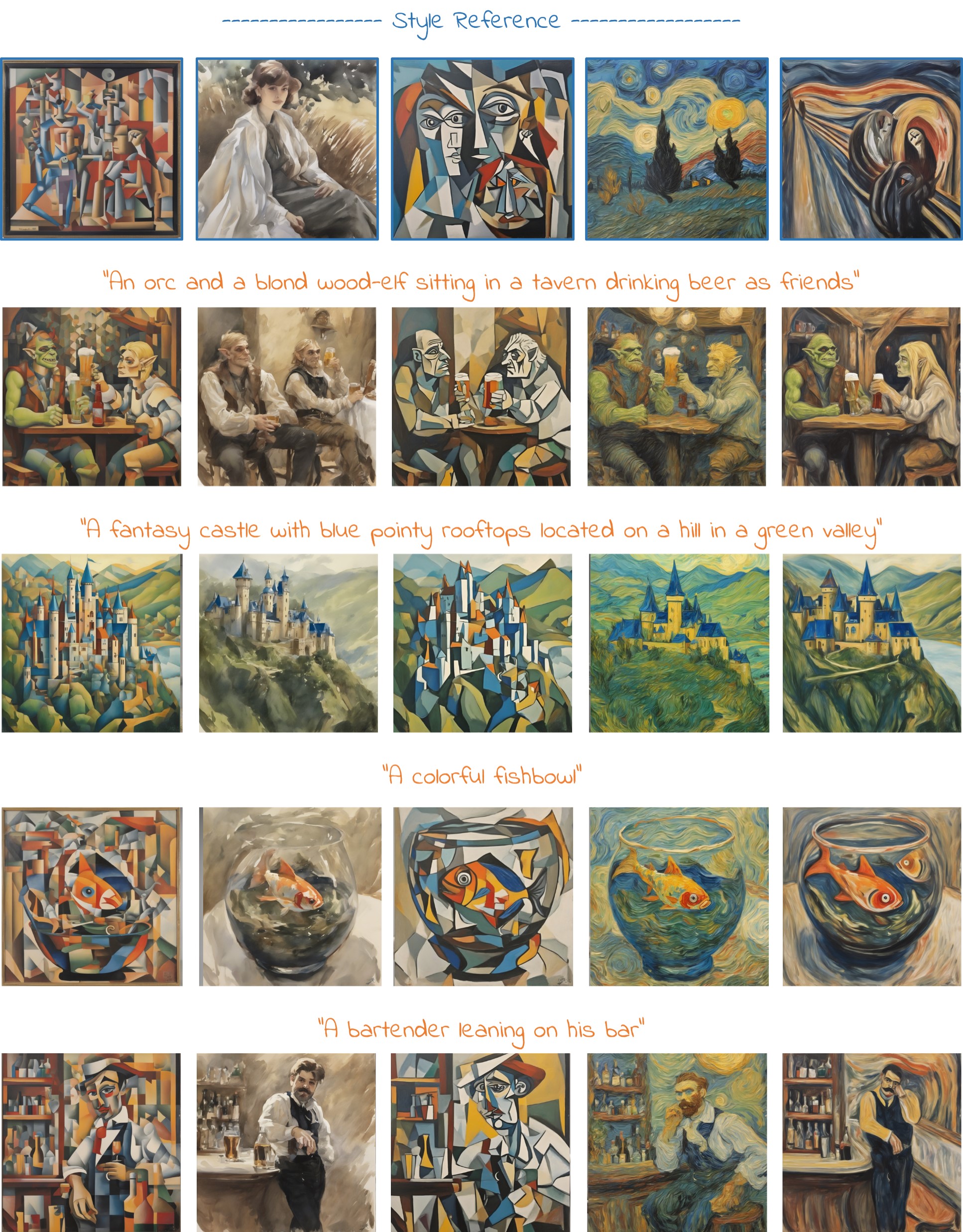}
    \caption{\textbf{Text Conditioned Results.} \textit{Zoom in for a better view.}}
    \label{fig:results1}
\end{figure*}

\begin{figure*}[t]  
    \centering
    \includegraphics[width=\textwidth]{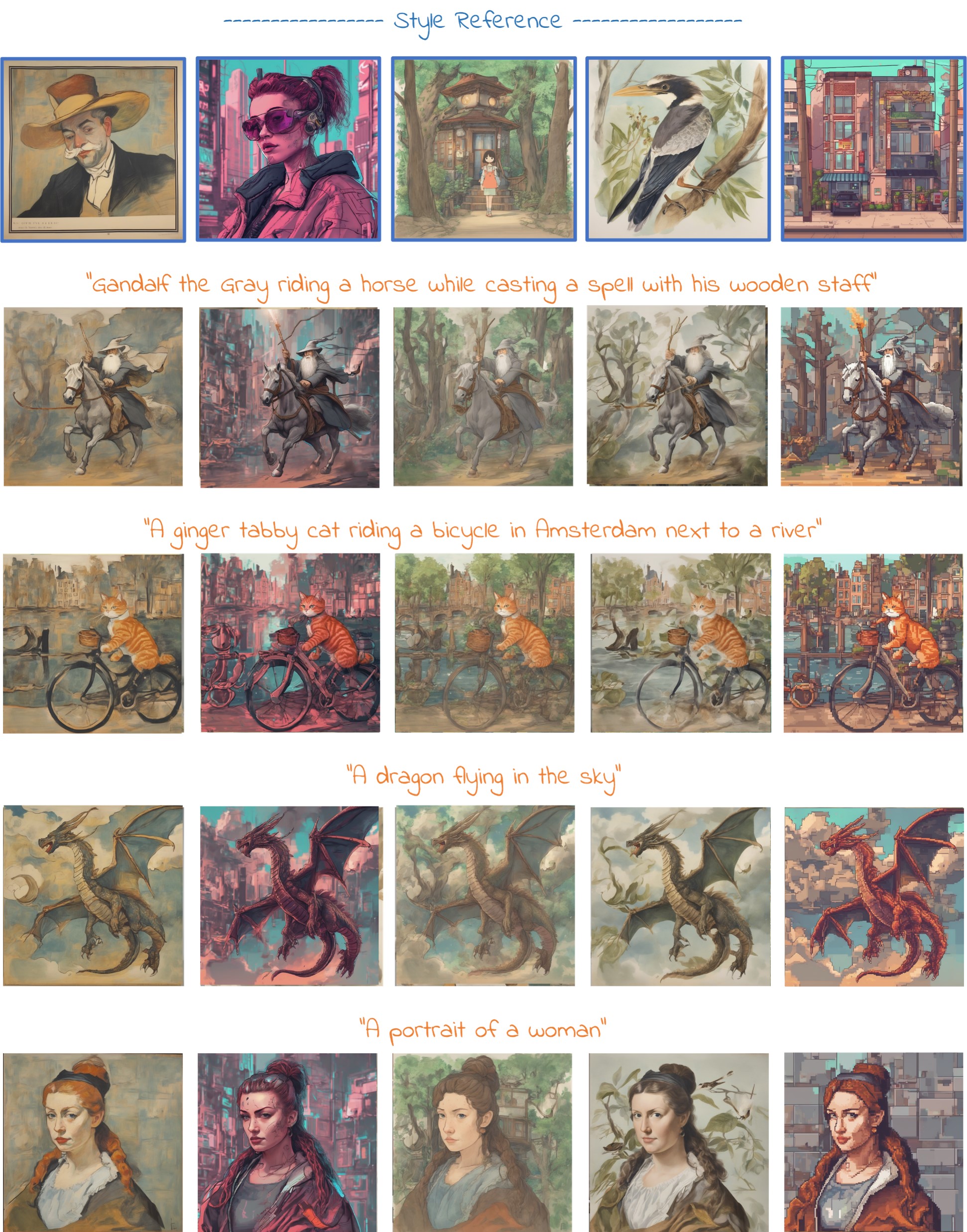}
    \caption{\textbf{Canny Conditioned Results.} \textit{Zoom in for a better view.}}
    \label{fig:results2}
\end{figure*}

\begin{figure*}[t]  
    \centering
    \includegraphics[width=\textwidth]{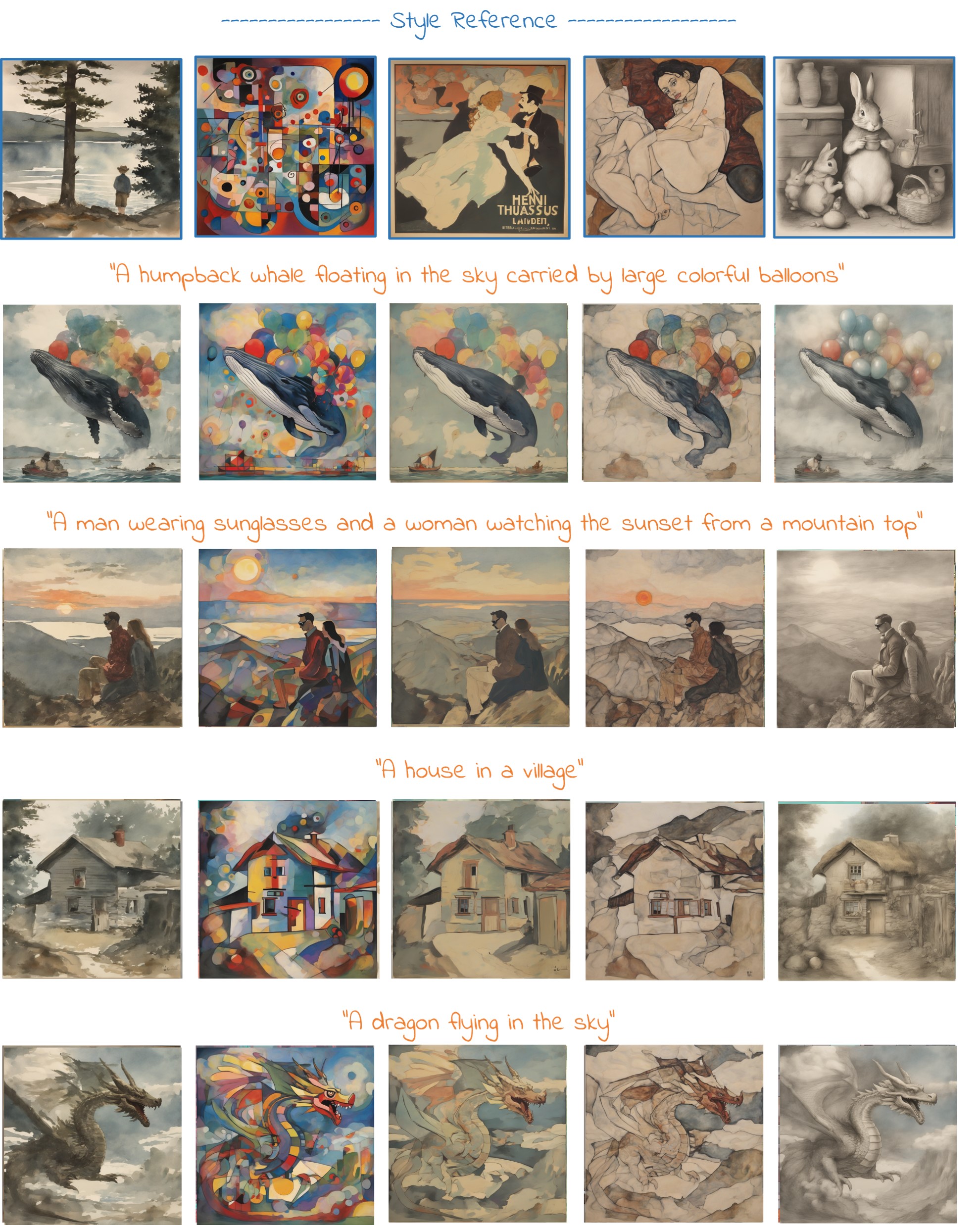}
    \caption{\textbf{Depth Conditioned Results.} \textit{Zoom in for a better view.}}
    \label{fig:results3}
\end{figure*}

\begin{figure*}[t]  
    \centering
    \includegraphics[width=\textwidth]{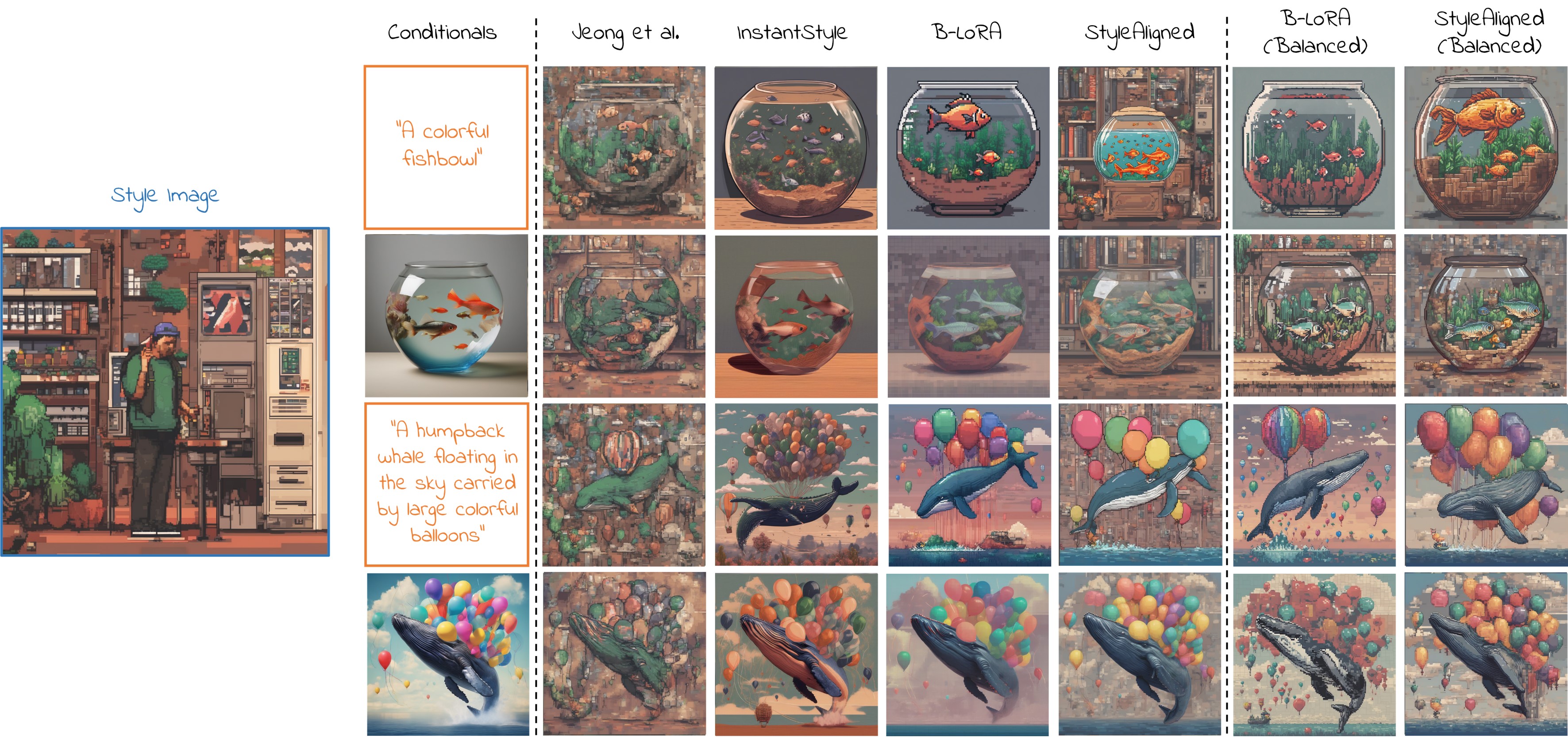}
    \caption{\textbf{Qualitative Comparison}. \textit{A comparison of different conditional combinations: Easy vs Complex prompt (two first rows vs. two last rows), Text only vs. Text and content image conditioning (1,3 vs 2,4 rows). As can be seen, both balanced methods achieves consistency over all conditioning combinations while the imbalanced methods show an inconsistent generation quality and in some examples content and style issues.}}
    \label{fig:comparison_whale}
\end{figure*}

\begin{figure*}[t]  
    \centering
    \includegraphics[width=\textwidth]{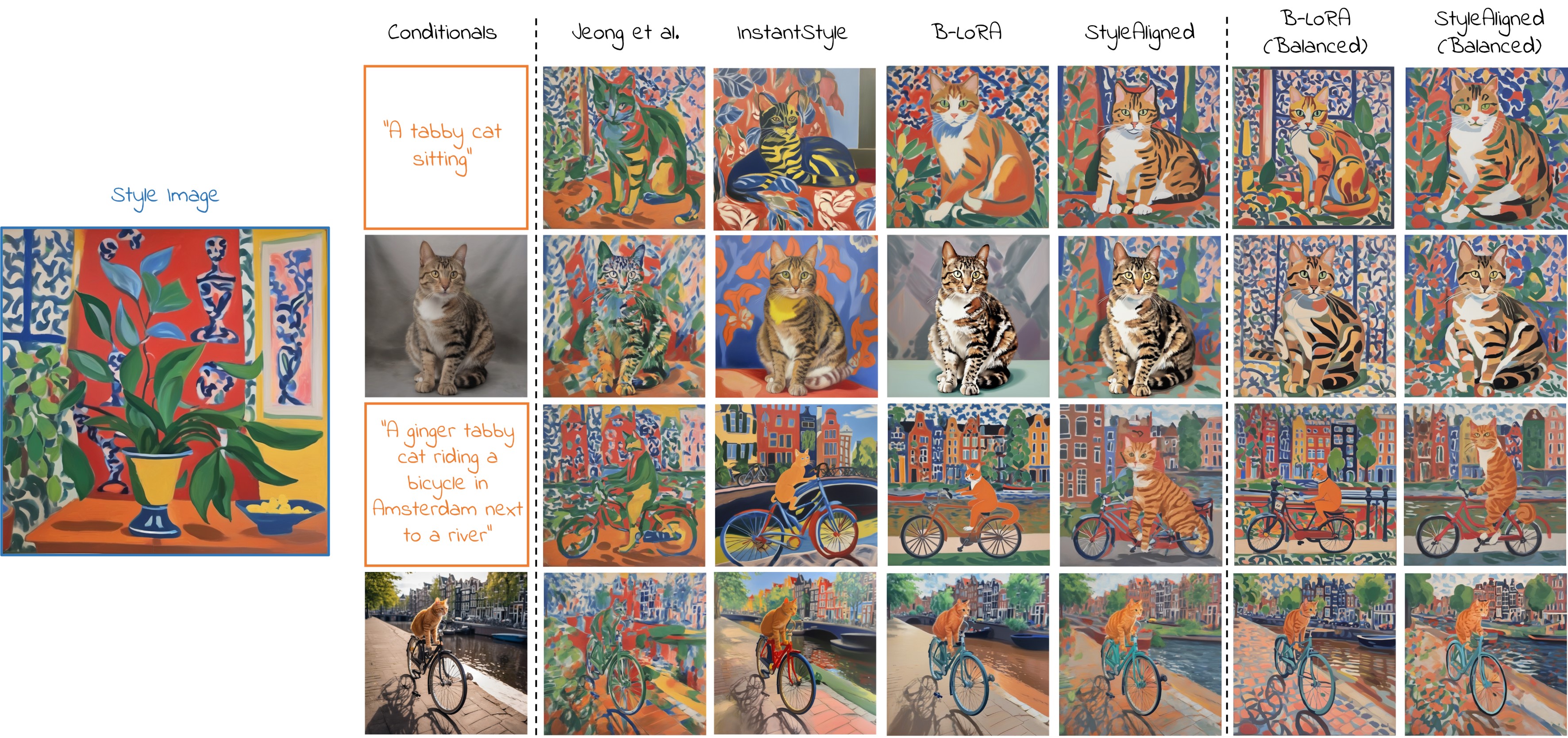}
    \caption{\textbf{Qualitative Comparison}. \textit{A comparison of different conditional combinations: Easy vs Complex prompt (two first rows vs. two last rows), Text only vs. Text and content image conditioning (1,3 vs 2,4 rows). As can be seen, both balanced methods achieves consistency over all conditioning combinations while the imbalanced methods show an inconsistent generation quality and in some examples content and style issues.}}
    \label{fig:comparison_cat}
\end{figure*}

\begin{figure*}[t]  
    \centering
    \includegraphics[width=\textwidth]{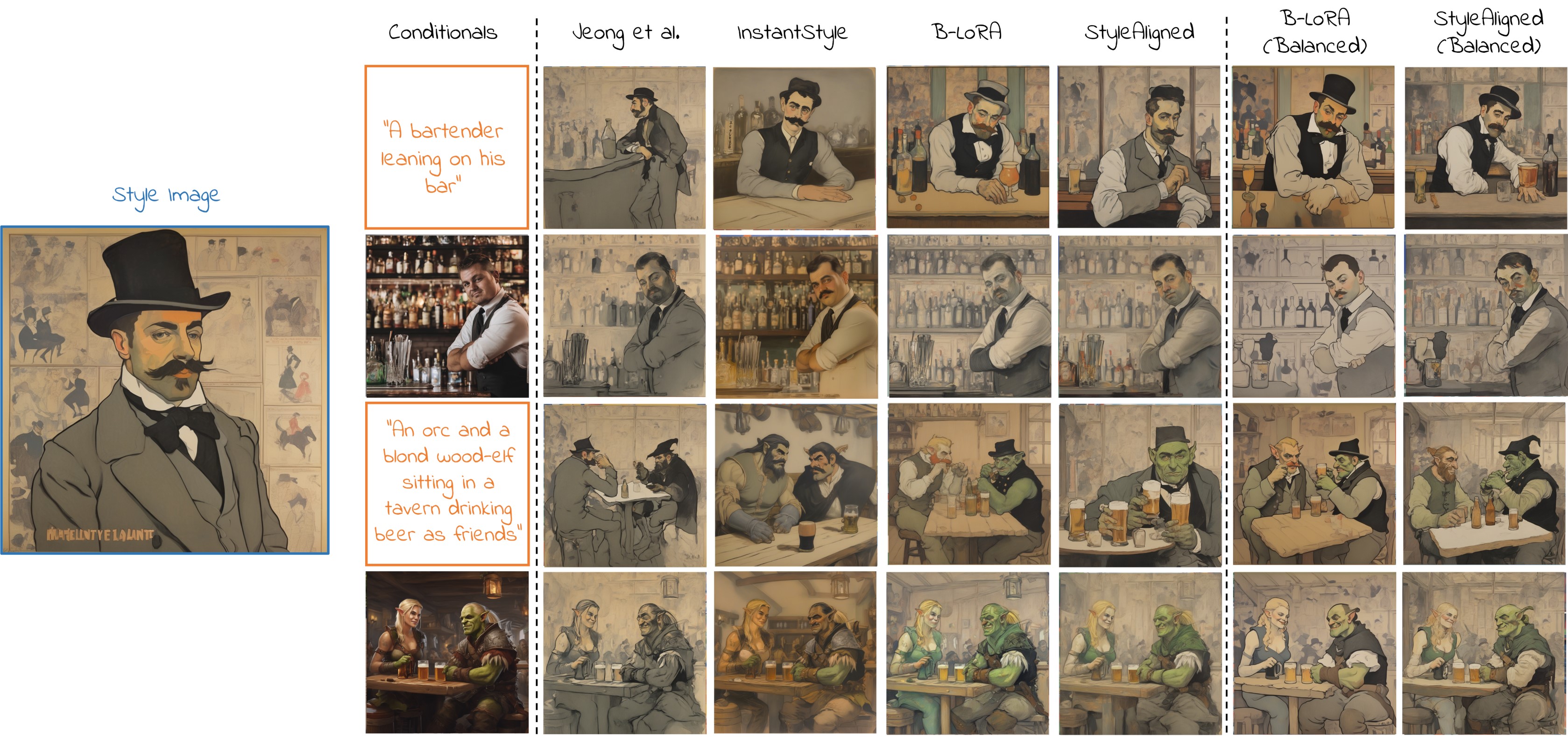}
    \caption{\textbf{Qualitative Comparison}. \textit{A comparison of different conditional combinations: Easy vs Complex prompt (two first rows vs. two last rows), Text only vs. Text and content image conditioning (1,3 vs 2,4 rows). As can be seen, both balanced methods achieves consistency over all conditioning combinations while the imbalanced methods show an inconsistent generation quality and in some examples content and style issues.}}
    \label{fig:comparison_bar}
\end{figure*}

\begin{figure*}[t]  
    \centering
    \includegraphics[width=\textwidth]{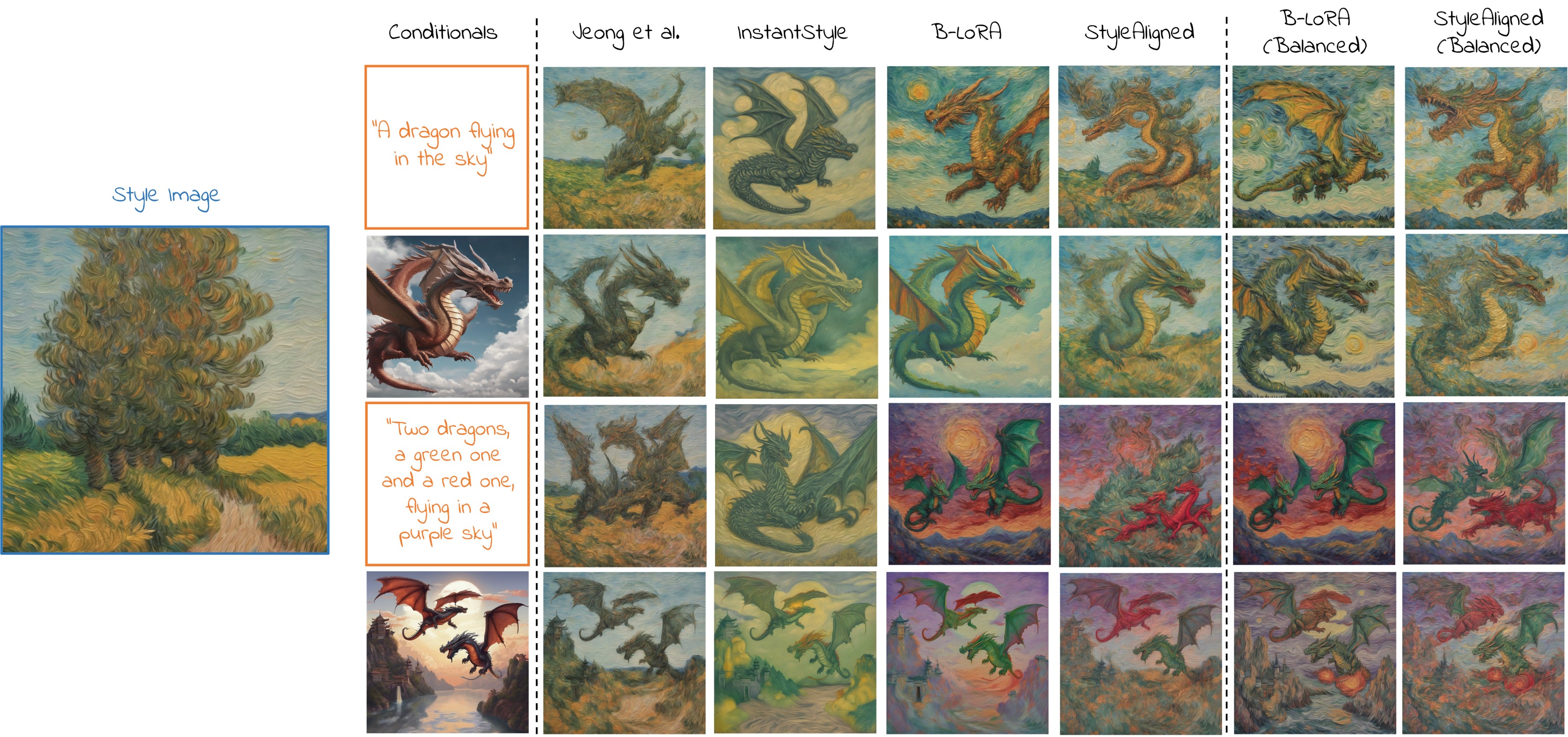}
    \caption{\textbf{Qualitative Comparison}. \textit{A comparison of different conditional combinations: Easy vs Complex prompt (two first rows vs. two last rows), Text only vs. Text and content image conditioning (1,3 vs 2,4 rows). As can be seen, both balanced methods achieves consistency over all conditioning combinations while the imbalanced methods show an inconsistent generation quality and in some examples content and style issues.}}
    \label{fig:comparison_dragon}
\end{figure*}

\begin{figure*}[t]  
    \centering
    \includegraphics[width=\textwidth]{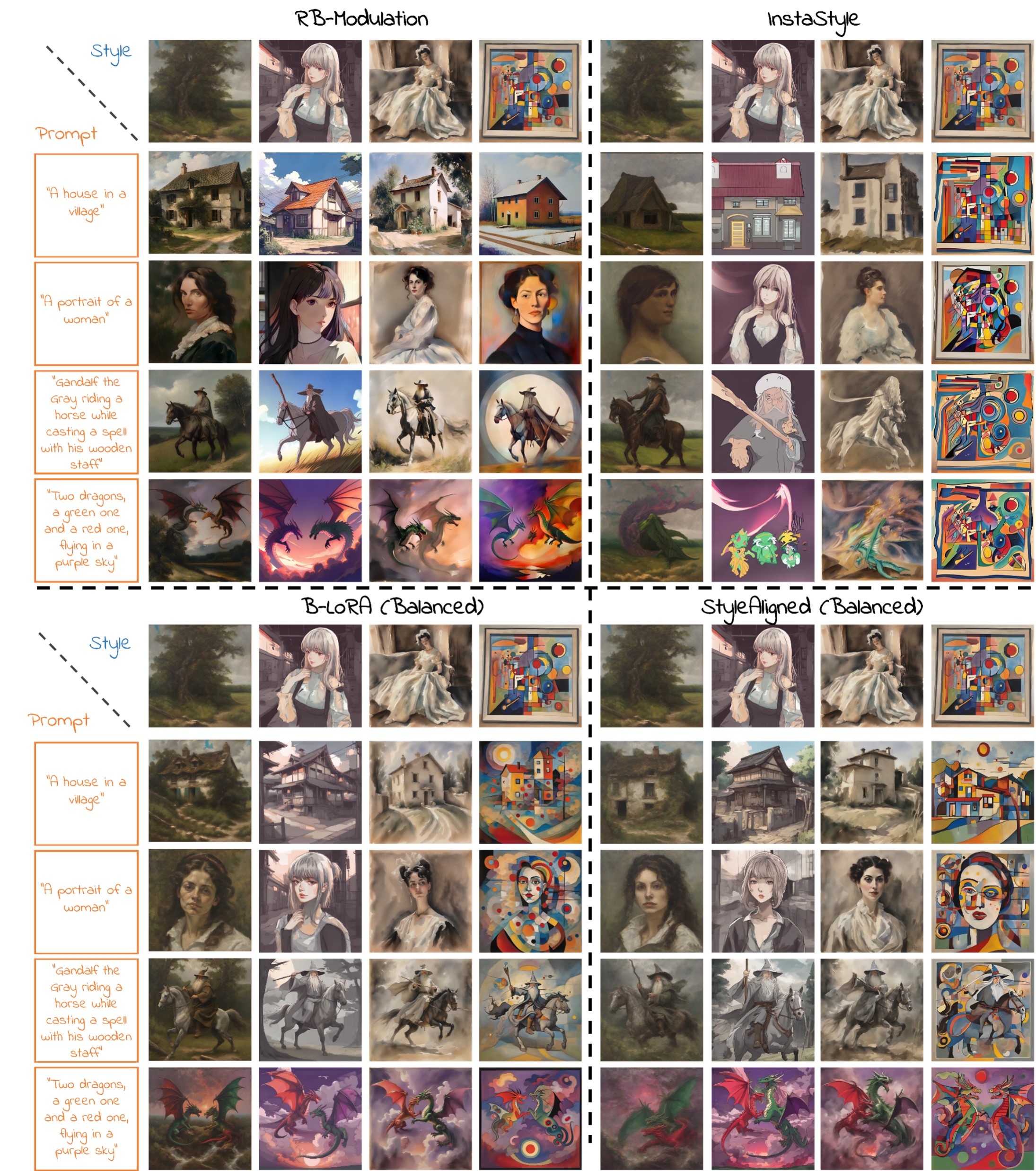}
    \caption{\textbf{Additional Comparisons}. \textit{Prompt + style image conditioned outputs for RB-Modulation (top left), InstaStyle (top right), Balanced B-LoRA (bottom left), and Balanced StyleAligned (bottom right.) Please zoom in for a better view.}}
    \label{fig:additional_comp}
\end{figure*}

\begin{figure*}[t]  
    \centering
    \includegraphics[width=\textwidth]{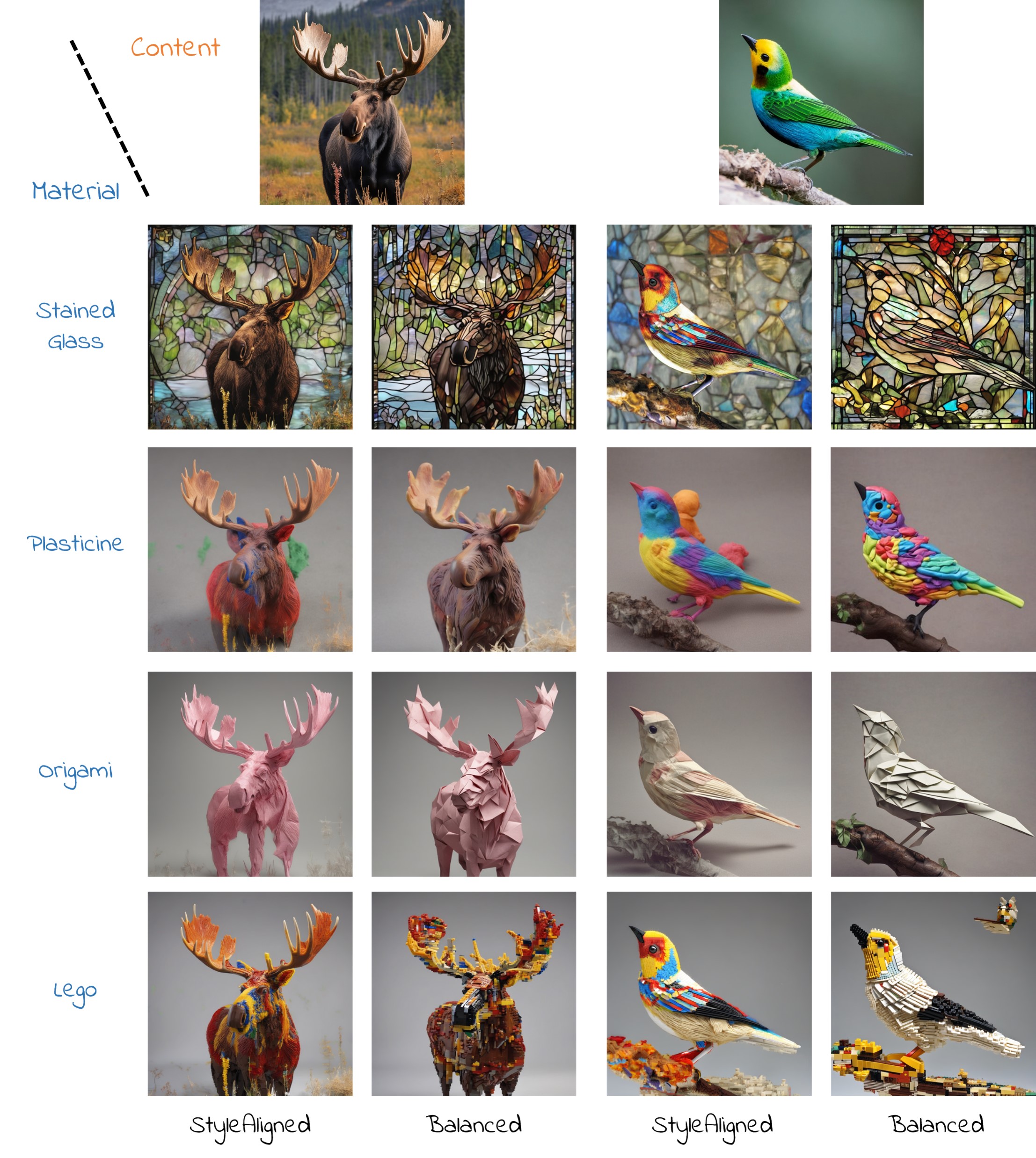}
    \caption{\textbf{Material Style Generation}. \textit{A sample of generated images with materialistic style, aligned to content images. Please zoom in for a better view.}}
    \label{fig:material1}
\end{figure*}

\begin{figure*}[t]  
    \centering
    \includegraphics[width=\textwidth]{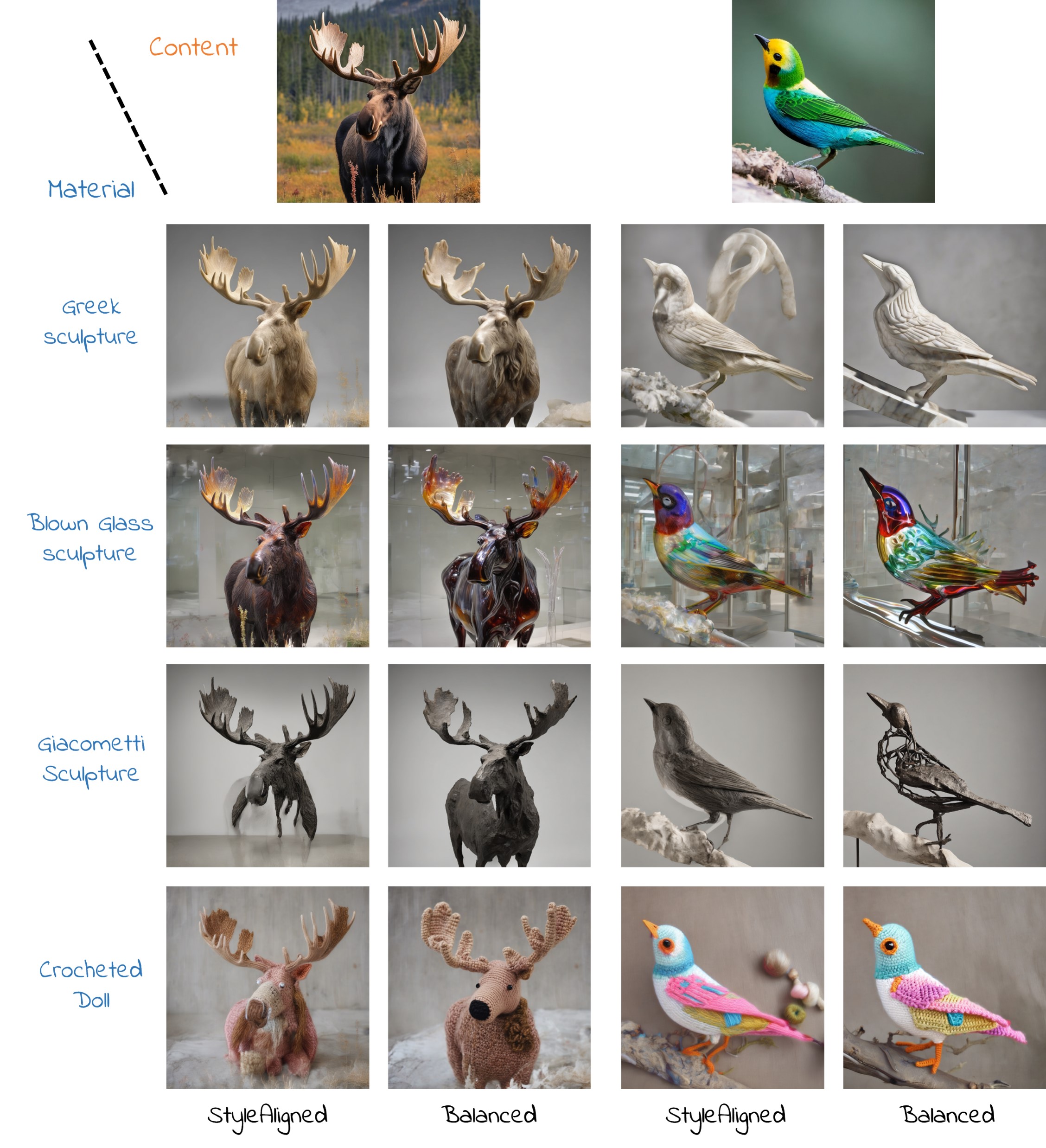}
    \caption{\textbf{Material Style Generation}. \textit{A sample of generated images with materialistic style, aligned to content images. Please zoom in for a better view.}}
    \label{fig:material2}
\end{figure*}

\begin{figure*}[t]  
    \centering
    \includegraphics[width=\textwidth]{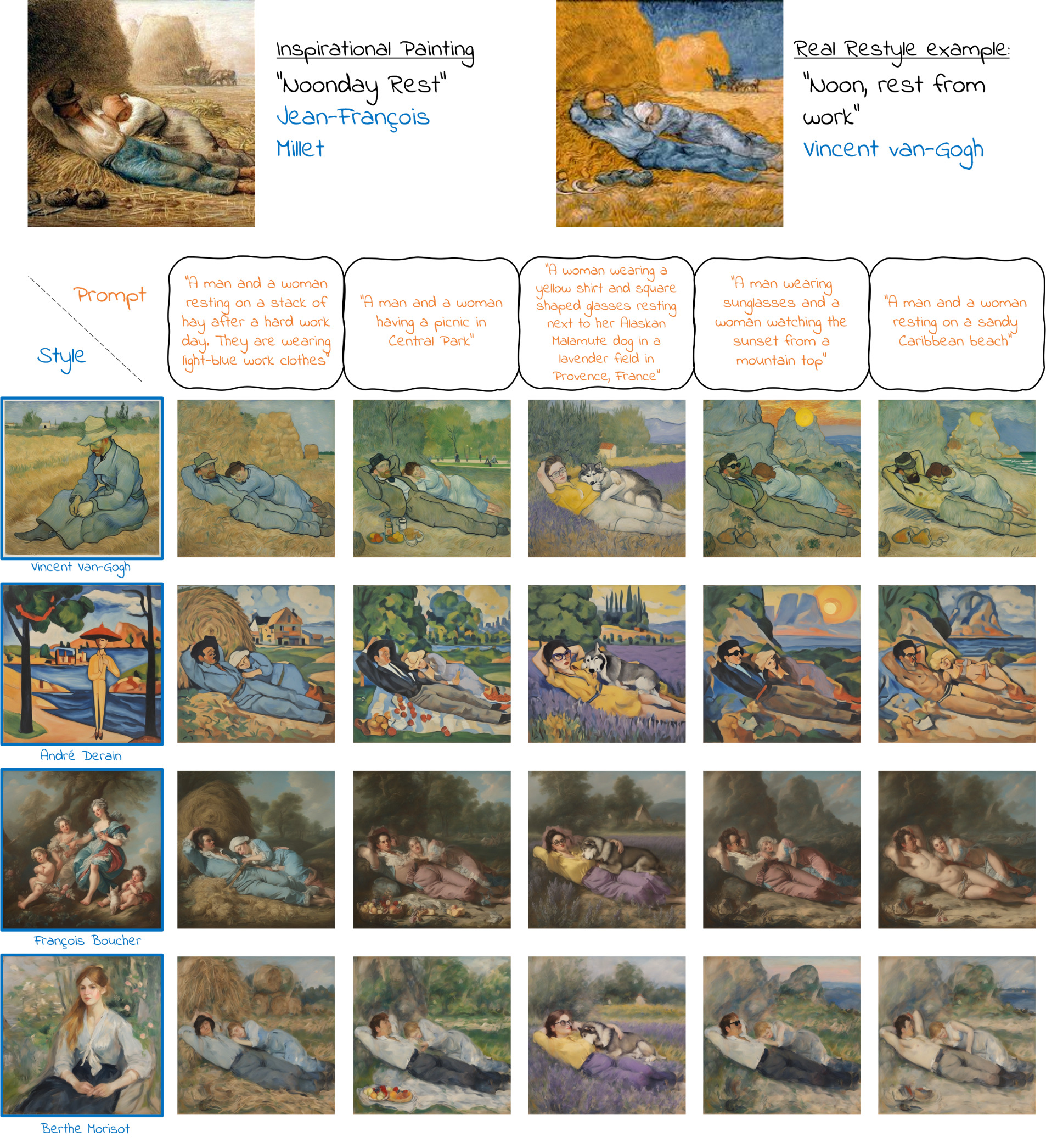}
    \caption{\textbf{Restyle/Recontent Example 1}. \textit{An example of restyling a painting inspired by van-Gogh's recreation of "Noonday Rest" by Jean-Francois Millet. The first column shows an example of restyling the content input without changing the original content while the rest of the columns shows an example of ReStyle and ReContent by editing both the image style and content of the output. (Please zoom in for a better view.)}}
    \label{fig:restyle_recontent1}
\end{figure*}

\begin{figure*}[t]  
    \centering
    \includegraphics[width=\textwidth]{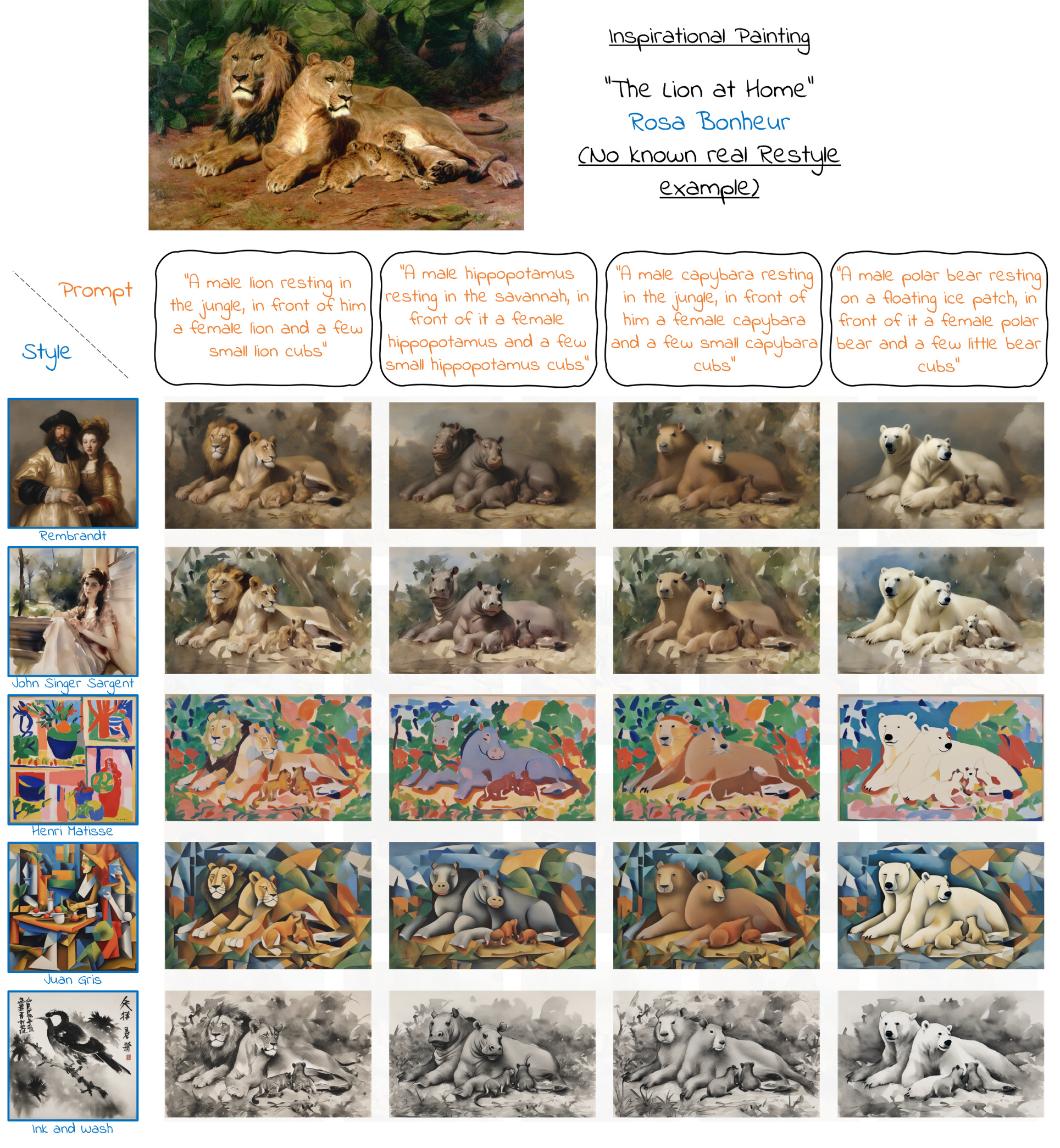}
    \caption{\textbf{Restyle/Recontent Example 2}. \textit{An example of restyling a painting of Rosa Bonheur: "The Lion at Home". The first column shows an example of restyling the content input without changing the original content while the rest of the columns shows an example of ReStyle and ReContent by editing both the image style and content of the output. (Please zoom in for a better view.)}}
    \label{fig:restyle_recontent2}
\end{figure*}

\begin{figure*}[t]  
    \centering
    \includegraphics[width=\textwidth]{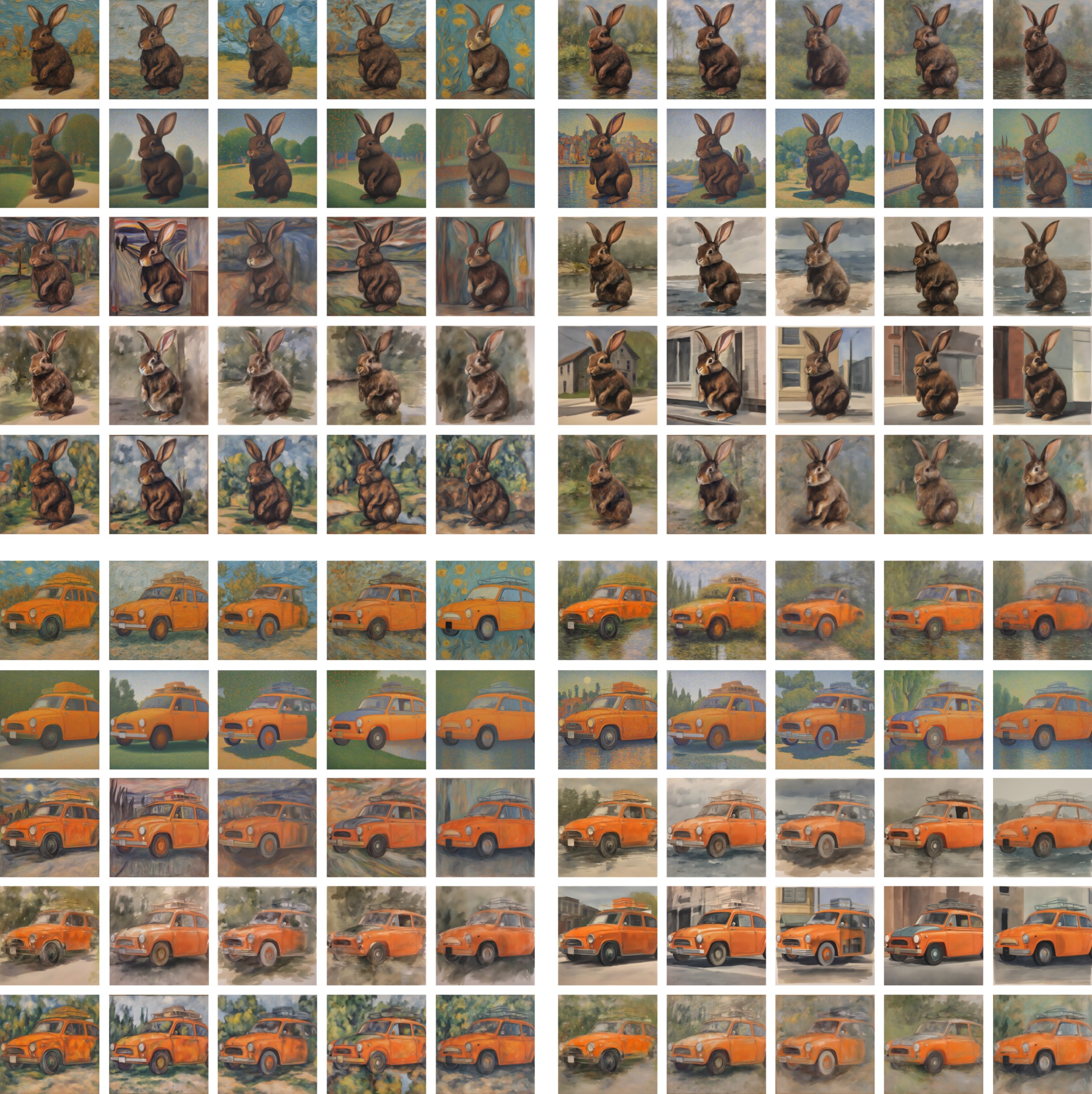}
    \caption{\textbf{Style Collections Example}. \textit{An example of a paintings collection used for our style sensitivity analysis.}}
    \label{fig:style_collection}
\end{figure*}

\begin{figure*}[t]  
    \centering
    \includegraphics[width=\textwidth]{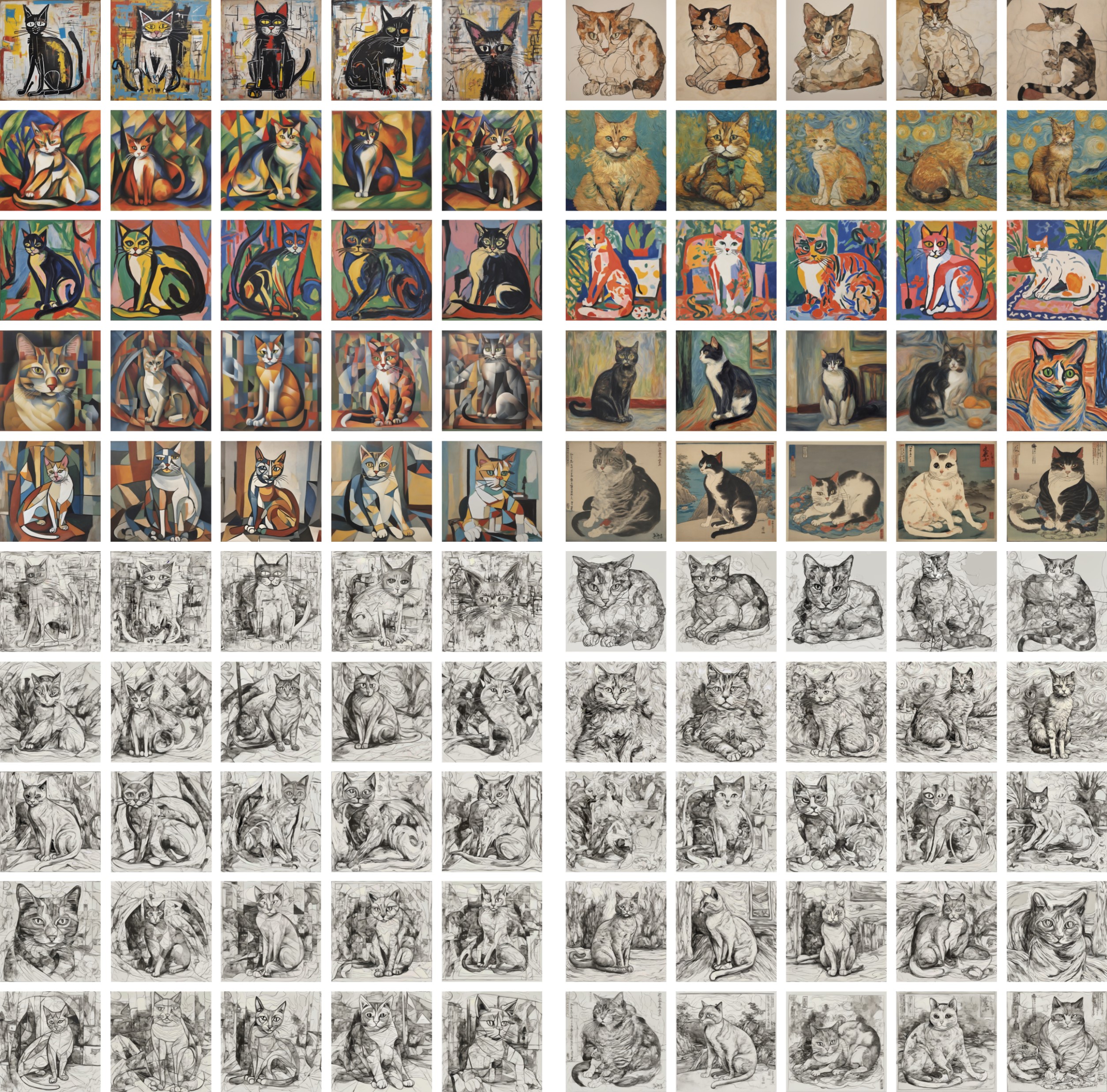}
    \caption{\textbf{Geometric Collection Example}. \textit{An example of a paintings collection used for our geometric style sensitivity analysis. Please zoom in for a better view.}}
    \label{fig:cat_geo_col}
\end{figure*}

\begin{figure*}[t]  
    \centering
    \includegraphics[width=\textwidth]{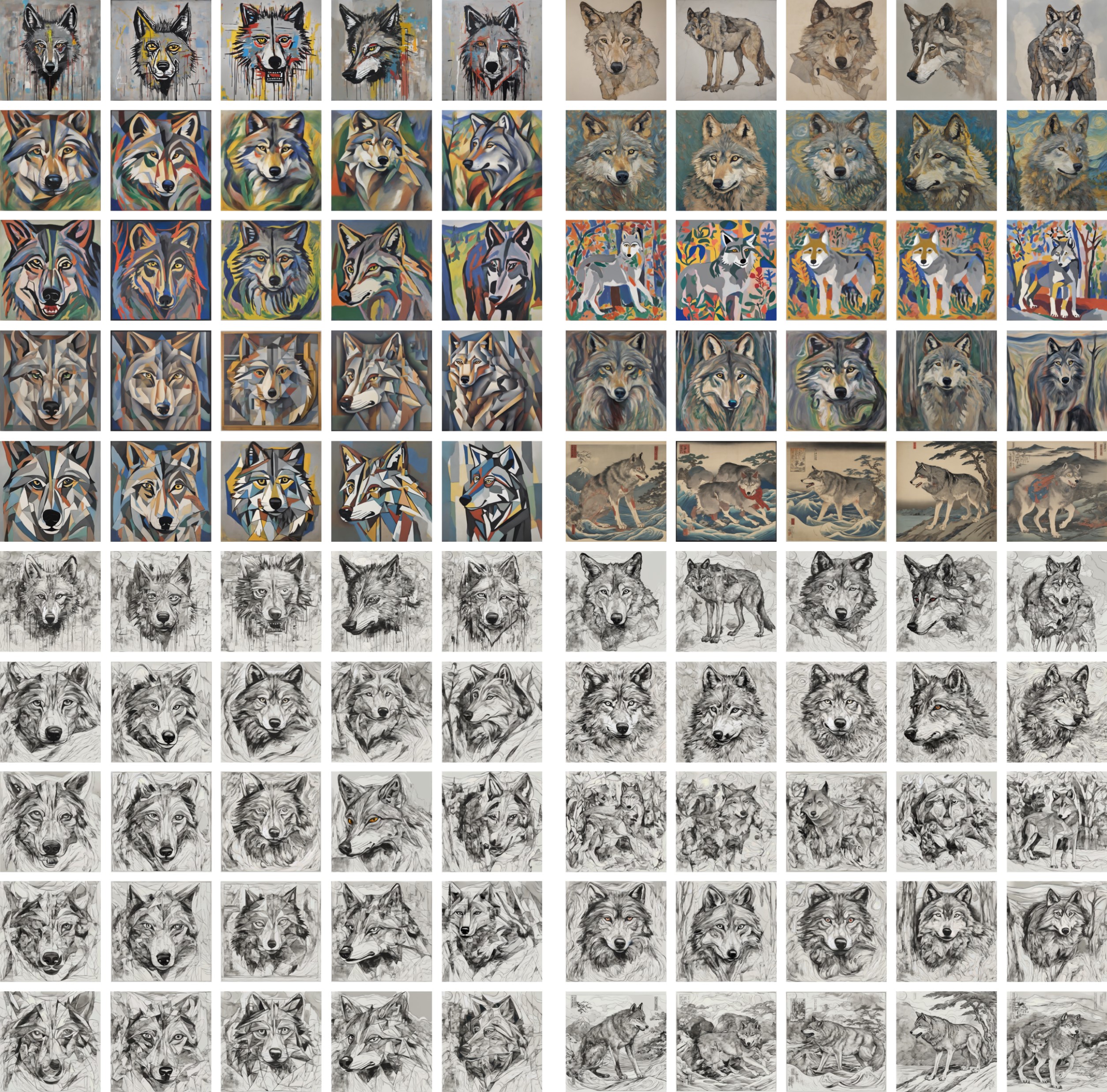}
    \caption{\textbf{Geometric Collection Example}. \textit{An example of a paintings collection used for our geometric style sensitivity analysis. Please zoom in for a better view.}}
    \label{fig:wold_geo_col}
\end{figure*}

\begin{figure*}[t]  
    \centering
    \includegraphics[width=\textwidth]{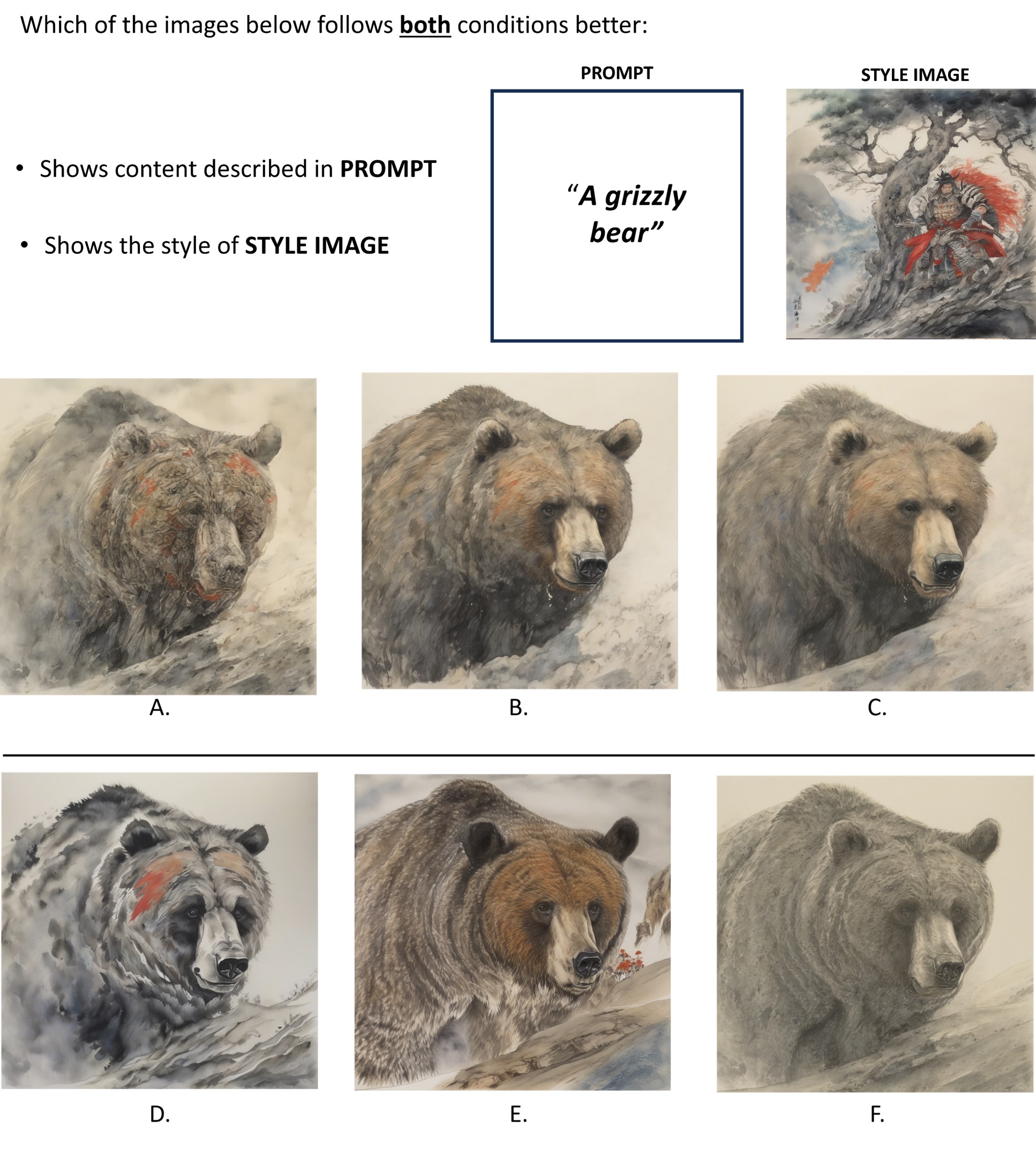}
    \caption{\textbf{User Study - Multiple Choice Questions.} \textit{A sample of a multiple choice question from the user study.}}
    \label{fig:rb_user}
\end{figure*}

\begin{figure*}[t]  
    \centering
    \includegraphics[width=\textwidth]{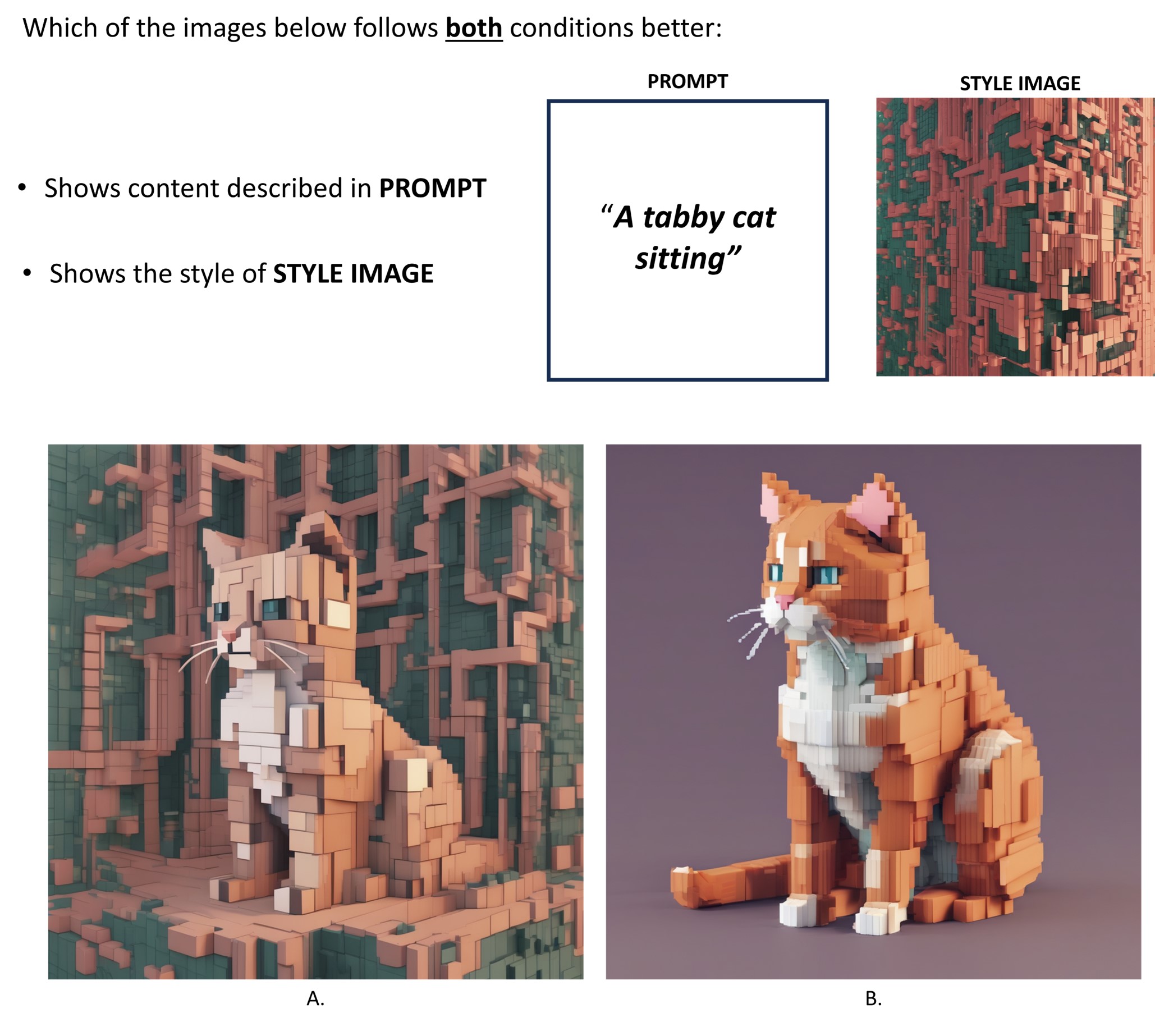}
    \caption{\textbf{User Study - A/B choice Questions.}. \textit{A sample of an A/B choice question from the user study.}}
    \label{fig:ab_user}
\end{figure*}